%% file: arxiv.tex
\documentclass[11pt]{article}

\date{}
\usepackage{times}

\usepackage[utf8]{inputenc} 
\usepackage[T1]{fontenc}    
\usepackage{hyperref}       
\usepackage{url}            
\usepackage{booktabs}       

\usepackage{amsfonts}       
\usepackage{amsmath}
\usepackage{amsthm}
\usepackage{amssymb}
\usepackage{mathtools}

\usepackage{nicefrac}       
\usepackage{microtype}      
\usepackage{xspace}
\usepackage{xcolor}
\usepackage{xfrac}

\usepackage{algorithm}
\usepackage[noend]{algpseudocode}

\usepackage[numbers, compress]{natbib}
\usepackage{caption}
\usepackage{enumitem}

\usepackage{graphicx}
\usepackage{float}

\input{define.tex}

\title{Active Symbolic Discovery of Ordinary Differential Equations via Phase Portrait Sketching}
\author{Nan Jiang$^1$, Md Nasim$^2$, Yexiang Xue$^1$ \\
$^1$Department of Computer Science, Purdue University, USA\\
    $^2$Department of Computer Science, Cornell University, USA \\
\texttt{ \{jiang631, yexiang\}@purdue.edu, md.nasim@cornell.edu}}

\begin{document}

\maketitle

\begin{abstract}
\input{tex/0-abstract}

\end{abstract}

\input{tex/1-intro}

\input{tex/2-prelim}

\input{tex/3.method}

\input{tex/4.related}

\input{tex/5.exp}
\input{tex/6.conclude}

\section*{Acknowledgments}
We thank all the reviewers for their constructive comments.
This research was supported by NSF grant CCF-1918327, NSF Career Award IIS-2339844, DOE – Fusion Energy Science grant: DE-SC0024583, and AI Climate: NSF and USDA-NIFA and  Cornell University AI for Science Institute.

\clearpage
\bibliography{reference}
\bibliographystyle{unsrtnat} 
\newpage

\appendix

\section{Availability of \method, Baselines and Dataset}
 Please find our code repository at:
\begin{mdframed}
\url{https://github.com/jiangnanhugo/ASD-ODE/}
\end{mdframed}
\begin{enumerate}
    \item the implementation of our \method method is in the folder ``\texttt{apps\_ode\_pytorch/}''.
    \item the list of datasets is listed in ``\texttt{data\_oracle/scibench/scibench/data/}''.
    \item the implementation of several baseline algorithms is collected in folder ``\texttt{baslines/}''.
\end{enumerate}
We provide a ``README.md'' document for executing the programs.

We summarize the supplementary material as follows: 
Section~\ref{apx:phase} offers a detailed explanation of the phase portrait for ODE and the concepts of active learning;
Section~\ref{apx:implement} provides the extended explanation of the proposed \method method; Section~\ref{apx:exp-set} details the experimental settings; Section~\ref{apx:extra-exp} collects the extra experimental result.

\input{tex/10.3.phase}

\input{tex/10.1.extend-method}

\input{tex/10.2.implement}

\input{tex/10.4.expset}

\end{document}

%% file: define.tex
\usepackage{enumitem}
\usepackage{bm}
\usepackage{amsmath,amssymb,amsfonts,amsthm}

\usepackage{graphicx}
\usepackage{booktabs}
\usepackage{multicol,multirow}
\usepackage{url}

\usepackage{xcolor}
\usepackage{xfrac}
\usepackage{xspace}

\usepackage{algorithm}
\usepackage[noend]{algpseudocode}
\usepackage{mdframed}

\newtheorem*{theorem-no}{Theorem}

\newtheorem*{lemma-no}{Lemma}

\theoremstyle{definition}

\algrenewcommand\algorithmicrequire{\textbf{Input:}}
\algrenewcommand\algorithmicensure{\textbf{Output:}}

\newcommand{\method}{\textsc{Apps}\xspace}

\newcommand{\dd}{\mathit{d}}

%% file: tex/0-abstract.tex
The symbolic discovery of Ordinary Differential Equations (ODEs) from trajectory data plays a pivotal role in AI-driven scientific discovery. Existing symbolic methods predominantly rely on fixed, pre-collected training datasets, which often result in suboptimal performance, as demonstrated in our case study in Figure~\ref{fig:time-seq}. 
Drawing inspiration from active learning, we investigate strategies to query informative trajectory data that can enhance the evaluation of predicted ODEs. However, the butterfly effect in dynamical systems reveals that small variations in initial conditions can lead to drastically different trajectories, necessitating the storage of vast quantities of trajectory data using conventional active learning.
To address this, we introduce \underline{\textbf{A}}ctive Symbolic Discovery of Ordinary Differential Equations via \underline{\textbf{P}}hase \underline{\textbf{P}}ortrait \underline{\textbf{S}}ketching (\method). Instead of directly selecting individual initial conditions, our \method first identifies an informative region within the phase space and then samples a batch of initial conditions from this region. 
Compared to traditional active learning methods, \method mitigates the gap of maintaining a large amount of data. Extensive experiments demonstrate that \method consistently discovers more accurate ODE expressions than baseline methods using passively collected datasets. 

%% file: tex/1-intro.tex
\section{Introduction}
Uncovering the governing principles of physical systems from experimental data is a crucial task in AI-driven scientific discovery~\cite{doi:10.1126/science.1165893,doi:10.1098/rspa.2018.0305,PhysRevE.100.033311}. Recent advancements have introduced various methods for uncovering knowledge of dynamical systems in symbolic Ordinary Differential Equation (ODE) form, leveraging techniques such as genetic programming~\cite{DBLP:conf/gecco/HeLYLW22}, sparse regression~\cite{brunton2016sparse,fasel2022ensemble}, Monte Carlo tree search~\cite{DBLP:conf/iclr/Sun0W023}, pretrained Transformers~\cite{DBLP:conf/iclr/QianKS22}, and deep reinforcement learning~\cite{DBLP:conf/ijcai/0012NX24}.

\begin{figure*}[!t]
    \centering
    \includegraphics[width=0.235\linewidth]{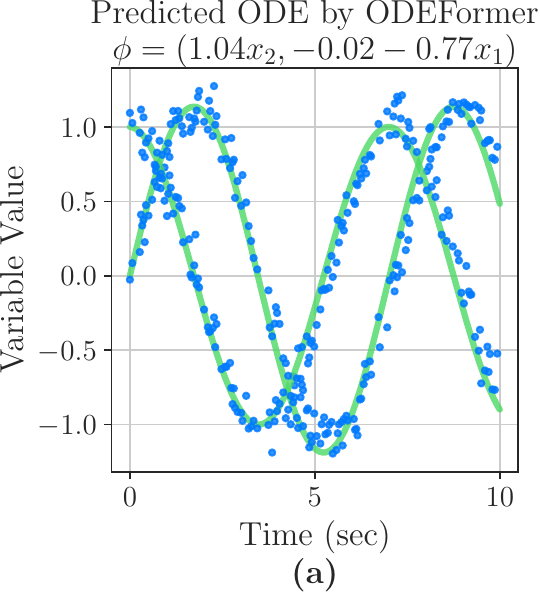}
    \includegraphics[width=0.215\linewidth]{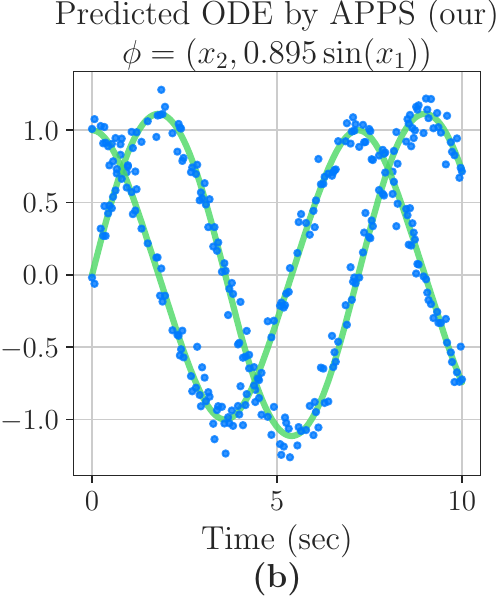}
    \hfill
    \includegraphics[width=0.215\linewidth]{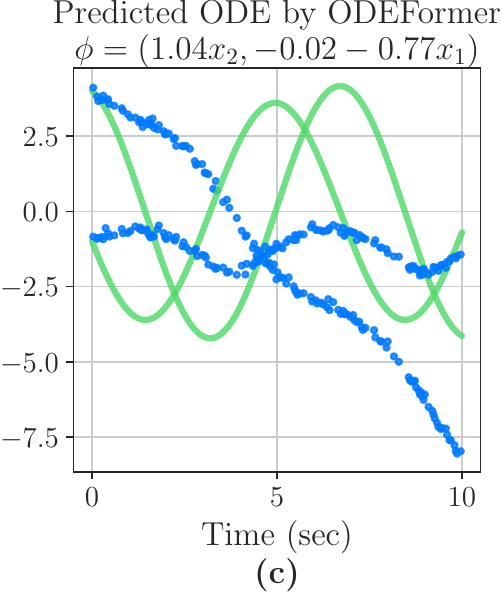}
     \includegraphics[width=0.319\linewidth]{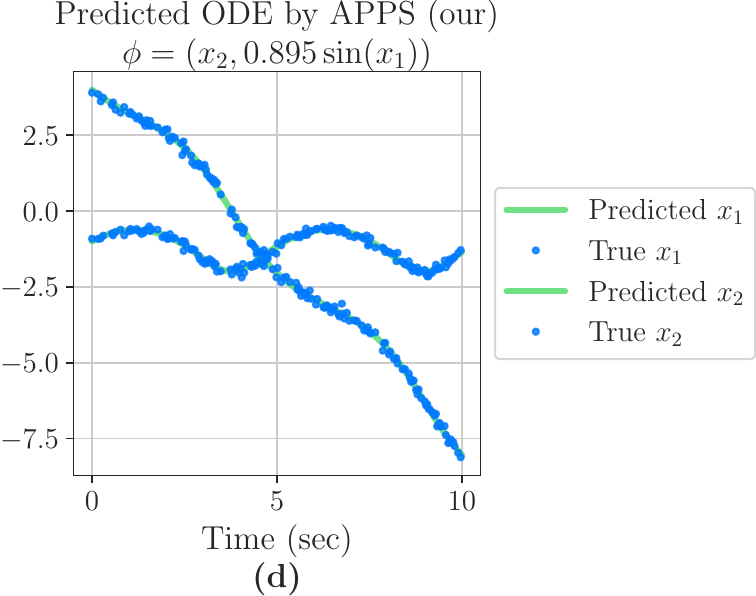}
    \caption{The performance of predicted ODE from passively-learned baseline is heavily influenced by the collected training data while our \method method is not.
    The dots represent noisy ground-truth trajectory data, and the lines show predicted values of state variables under identical initial conditions. \textbf{(a, b)}  Our \method and the baseline predict accurately for the trajectory starting at $\mathbf{x}_0=(0,1)$. \textbf{(c, d)} For the trajectory starting at $\mathbf{x}_0=(4,-1)$, the baseline performs poorly while \method maintains accuracy.}
    \label{fig:time-seq}
\end{figure*}

State-of-the-art approaches discover the symbolic ODEs using a fixed, pre-collected training dataset. However, their performance is often heavily influenced by the quality of the collected data. As illustrated in Figure~\ref{fig:time-seq}, we find that the best-discovered ODEs from the most recent baseline, that is ODEFormer~\cite{2023odeformer}, may fit some test trajectories well, but fit other test trajectories poorly.
This observation highlights the need for new methods that actively query informative trajectory data to improve ODE discovery. 

Suppose trajectory data can be obtained from a data oracle by specifying the initial conditions. To minimize excessively querying the oracle, a key challenge emerges: given a set of candidate ODEs predicted by a learning method, how can initial conditions within the variable intervals be strategically selected to obtain informative data?

Previous work in the active learning literature typically maintains a large set of data, evaluates their informativeness, and then queries the most informative data points~\cite{golovin2010near,medina2023active}. However, the chaotic nature of dynamical systems complicates the direct application of such methods. The Butterfly effect states that small variations in initial conditions can lead to vastly different outcomes. For instance, as illustrated in Figure~\ref{fig:pipeline}(c), selecting initial conditions near $(3, 0)$ for $\phi_1$ can result in trajectories that diverge in opposite directions. Effectively addressing this variability requires densely sampling initial conditions to thoroughly explore the space. Existing active learning-based approaches will be computationally prohibitive and demand significant memory resources, particularly in high-dimensional dynamical systems.

To address these challenges, we propose a novel approach to data querying. We consider selecting a batch of close-neighbor initial conditions instead of individual initial conditions. This process begins by sketching the dynamics in smaller regions, identifying an \textit{informative region} in the phase space, and then sampling a batch of initial conditions from this region. Figure~\ref{fig:pipeline}(c) illustrates this idea using phase portraits for three candidate ODEs. Region $u_2$ is chosen because the trajectories generated by the candidate ODEs exhibit greater divergence in this region than region $u_1$. 
Section~\ref{sec:method} provides detailed region selection criteria.

Thus, we introduce \underline{\textbf{A}}ctive Symbolic Discovery of Ordinary Differential Equations via \underline{\textbf{P}}hase \underline{\textbf{P}}ortrait \underline{\textbf{S}}ketching (\method), which consists of two key components: (1) a deep sequential decoder, which guides the search for candidate ODEs by sampling from the defined grammar rules. (2) a data query and evaluation module that actively queries the data using sketched phase portraits and evaluates the candidate ODE.  In experiments, we evaluate \method against several popular baselines on two large-scale ODE datasets. 
1) \method achieves the lowest median NMSE (in Table~\ref{tab:nmse} and Table~\ref{tab:nmse-extend}) across multiple datasets under noiseless and noisy settings.
2) Compared to other active learning strategies, \method is more time efficient in benchmark datasets (in Table~\ref{tab:diff-active}). \footnote{The code is at \url{https://github.com/jiangnanhugo/APPS-ODE}. Please refer to \url{https://arxiv.org/abs/2409.01416} for the appendix.}.

%% file: tex/2-prelim.tex
\section{Preliminaries} \label{sec:prelim}
    
\noindent\textbf{Ordinary Differential Equations} (ODEs) describe the evolution of dynamical systems in continuous time.
Let vector $\mathbf{x}(t)=(x_1(t),\ldots,x_n(t))\in\mathbb{R}^n$ be the state variables of the system of time $t$. 
    The temporal evolution of the system is governed by the time derivatives of the state variables, denoted as $  \frac{\dd x_i}{\dd t}$.
    The general form of the ODE is written as:
    \begin{equation*}
     \frac{\dd x_i}{\dd t} =f_i(\mathbf{x}(t),\mathbf{c}), \quad \text{for }  i=1,\ldots, n, 
    \end{equation*}
    where $f_i$ can be a linear or nonlinear function of the state variables $\mathbf{x}$ and coefficients $\mathbf{c}$.
    The ODE is noted as a tuple $(f_1,f_2,\ldots, f_n)$ for simplicity in this paper. 
    Function $f_i$ is symbolically expressed using a subset of input variables in $\mathbf{x}$ and coefficients in $\mathbf{c}$, connected by mathematical operators such as addition and cosine functions.  For example, we use $(10 \sin(x_2), 4 \cos(x_1 + 2))$ to represent the ODE $\frac{\dd x_1}{\dd t}=10 \sin(x_2), \frac{\dd x_2}{\dd t}=4\cos(x_1+2)$.

    Given an initial condition $\mathbf{x}_0$, the solution to the ODE is a \textit{trajectory} of state variables $(\mathbf{x}_0, \mathbf{x}(t_1), \ldots, \mathbf{x}(t_k))$ observed at discrete time points $(t_1, \ldots, t_k)$, possibly with noise. The trajectory is noted as $\tau$ for simplicity.

\begin{figure*}[!t]
    \centering
    \includegraphics[width=1\linewidth]{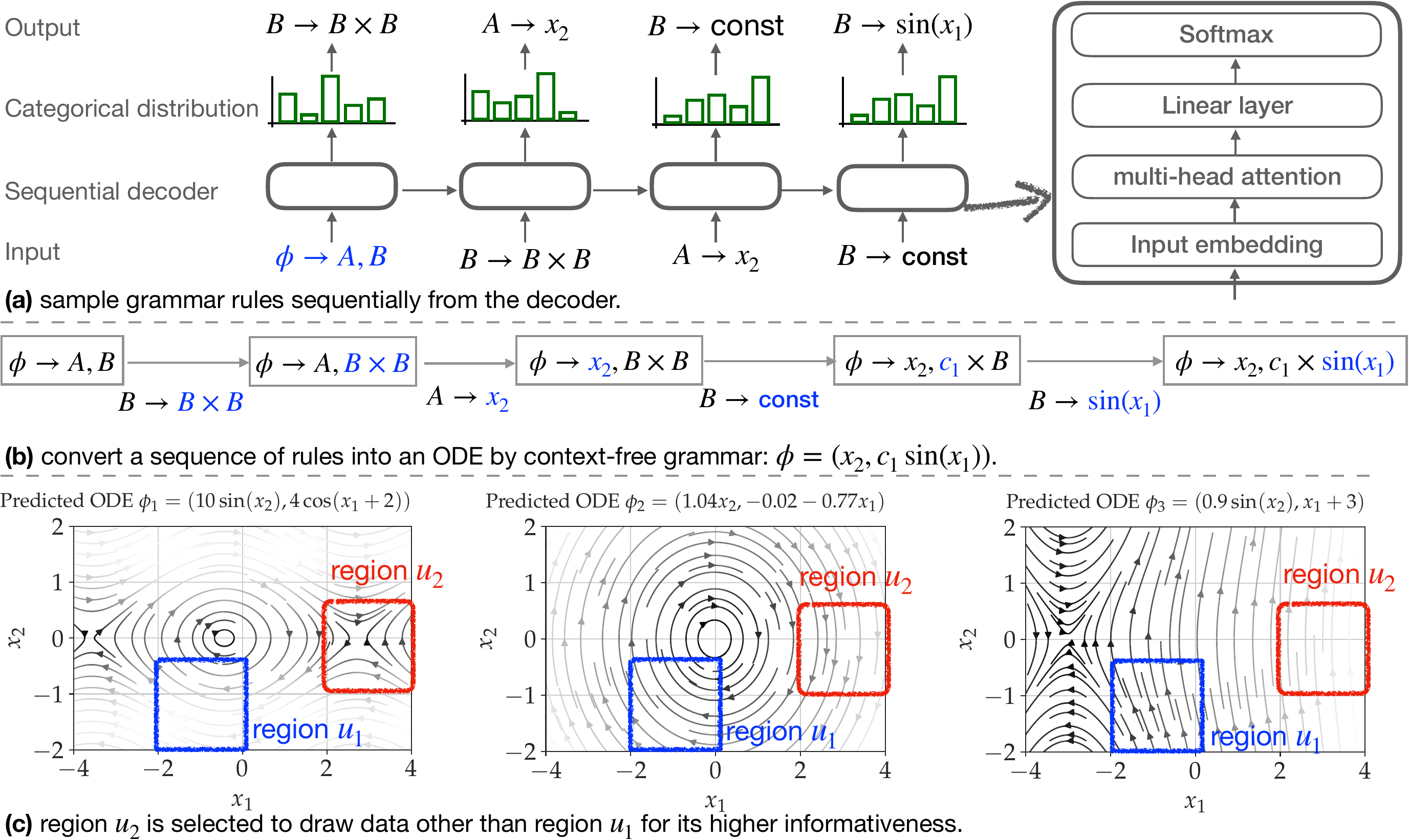}
    \caption{The pipeline of \method for symbolic discovery of ODEs consists of 3 steps: \textbf{(a)} ODEs are sampled from the sequential decoder by iteratively sampling grammar rules. The predicted rule at each step serves as input for the decoder in the subsequent step. \textbf{(b)} The sampled sequence of grammar rules is converted into a valid ODE with $n=2$ variables. Each rule expands the first non-terminal symbol, with the expanded parts highlighted in blue colors for clarity. \textbf{(c)} The phase portrait for the predicted ODEs (e.g., $\phi_1, \phi_2, \phi_3$) is sketched, and regions with high informativeness, such as $u_2$, are identified to query the new trajectory data. In region $u_2$, $\phi_1$ exhibits a saddle point, $\phi_2$ moves downward, and $\phi_3$ moves upward. In contrast, in region $u_1$, all trajectories move from right to left. Differentiating the predicted expressions is easier in region $u_2$ than in region $u_1$.}
    \label{fig:pipeline}
\end{figure*}

 \noindent\textbf{Phase Portrait} is a qualitative analysis tool for studying the behavior of dynamical systems~\cite{strogatz2018nonlinear}. Phase portraits are plotted using the state variables $\mathbf{x}$ and their time derivatives $(f_1,\ldots, f_n)$. A curve in the phase portrait is a short trajectory of the system over time from a given initial condition. The arrow on the curve indicates the direction of change. By examining these curves, we can infer key properties of the system, such as stability, equilibrium points, and periodic behavior. Figure~\ref{fig:pipeline}(c) shows phase portraits for three different ODEs. These portraits are generated by sampling random initial conditions within the variable intervals and evolving the system for a short time.

\noindent\textbf{Symbolic Discovery of Ordinary Differential Equations} seeks to uncover the symbolic form of an ODE that best fits a dataset of observed trajectories. 
According to~\citet{DBLP:conf/dis/GecOBDT22} and \citet{DBLP:conf/iclr/Sun0W023}, 
we are given a dataset of collected trajectories $D=\{\tau_1,\ldots, \tau_N\}$ and a set of mathematical operators $\{+,-,\times,\div,\sin\ldots\}$.
Denote $\phi(\mathbf{x}(t),\mathbf{c})$ as a candidate ODE, where $\mathbf{c}$ indicates the coefficients.  
The objective is to predict the symbolic form of the ODE that minimizes the distance between the predicted and observed trajectories, which is formalized as an optimization problem:
\begin{equation} \label{eq:obj}
\begin{aligned}
    \arg\min_{\phi \in \Pi} \frac{1}{|D|} &\sum_{\tau\in D} \sum_{i=1}^k \ell(\mathbf{x}(t_i), 
    \hat{\mathbf{x}}(t_i)), \\
    \text{where}&\qquad\hat{\mathbf{x}}(t_i) = {\mathbf{x}}_0 + \int_0^{t_i} \phi(\mathbf{x}(t),\mathbf{c})\dd t.
\end{aligned}
\end{equation}
$\Pi$ is the set of all possible ODEs, trajectory $\tau :=(\mathbf{x}_0, \mathbf{x}(t_1)$, $ \ldots, \mathbf{x}(t_k))$, $\mathbf{x}(t)$ is the ground-truth observations of the state variable. Trajectory $(\mathbf{x}_0, \hat{\mathbf{x}}(t_1)$, $ \ldots, \hat{\mathbf{x}}(t_k))$ is the predicted state variables according to the candidate ODE $\phi$.
The predicted trajectory $(\mathbf{x}_0, \hat{\mathbf{x}}(t_1), \ldots, \hat{\mathbf{x}}(t_k))$ is obtained by numerically integrating the ODE from the given initial state $\mathbf{x}_0$ to the final time $t_k$. The loss function $\ell$ computes the summarized distance between the predicted and ground-truth trajectories at each time step. A typical loss is the Normalized Mean Squared Error (NMSE, defined in Appendix D). Except for the above formulation, prior works in symbolic regression use the approximated time derivative as the \textit{label} to discover each expression $f_i$ separately, which is known as gradient matching. We leave the discussion to Related Work.
    
Recent research explored deep reinforcement learning to discover the governing equations from data~\cite{DBLP:conf/iclr/PetersenLMSKK21, DBLP:journals/corr/abs-1801-03526, DBLP:conf/nips/MundhenkLGSFP21}. In these approaches, each expression is represented as a binary tree, with interior nodes corresponding to mathematical operators and leaf nodes to variables or constants. An ODE with $n$ variables is represented by $n$ trees. The key idea is to frame the search for different ODEs as a sequential decision-making process based on the preorder traversal sequence of expression trees. A high reward is assigned to candidates which fit the data well. The search is guided by a deep sequential decoder, often based on RNN, LSTM, or decoder-only Transformer, that learns the optimal probability distribution for selecting the next node in the expression tree at each step. The parameters of the decoder are trained with the policy gradient algorithm.

%% file: tex/3.method.tex
\section{Methodology} \label{sec:method}

\subsection{Motivation}
For the task of symbolic discovery of ODEs, we observe that existing methods frequently overfit the training data. This issue is illustrated in Figure~\ref{fig:time-seq} using ODEFormer~\cite{2023odeformer}, a recent baseline designed to learn ODEs from a fixed training dataset. In the example, the best-predicted ODE is given by $\phi = (1.04x_2, -0.02 - 0.77x_1)$. We evaluate $\phi$ on noisy test trajectories (depicted as blue dots) with two distinct initial conditions. While $\phi$ closely aligns with the trajectory originating at $\mathbf{x}_0 = (0, 1)$, as shown by the green curve, it produces substantial errors for a trajectory starting at $\mathbf{x}_0 = (4, -1)$, where the predicted curve deviates significantly from the ground truth.

This observation motivates us to actively identify informative trajectory data to better differentiate candidate expressions during the learning process. Each trajectory is generated by querying the data oracle with a specified initial condition $\mathbf{x}_0$. An initial condition is deemed \textit{informative} if the resulting trajectory for different candidate ODEs diverges significantly. The key challenge lies in selecting such informative initial conditions from the variable intervals for a given set of candidate ODEs.

For addressing this issue, a common approach in active learning~\cite{golovin2010near} is to maintain a large set of potential initial conditions, evaluate their informativeness, and query the most informative points. However, the butterfly effect in chaos theory~\cite{lorenz1963deterministic} suggests existing works in active learning are not directly applicable. The chaotic behavior states small changes in initial conditions can lead to drastically different outcomes in dynamical systems. For example, as shown in Figure~\ref{fig:pipeline}(c), selecting points near $(3, 0)$ (inside the red region $u_2$) for $\phi_1$ can lead to trajectories diverging either towards the top right or the bottom left. 
Such chaotic behavior necessitates the existing active learning methods to maintain a large set of initial conditions to adequately cover the domain, which becomes infeasible for high-dimensional dynamical systems.

To mitigate this issue, we consider selecting a beam of near-neighbor points rather than individual points. We propose first to select a highly informative region and sample a batch of initial conditions within that region. In this research, the region is represented as an $n$-dimensional cube of fixed width. A region is regarded as \textit{informative} if the majority of sampled initial conditions within it yield informative trajectories for the given candidate ODEs.

Figure~\ref{fig:pipeline}(c) illustrates our region-based approach using the phase portraits of three candidate ODEs: $\phi_1, \phi_2$, and $\phi_3$. Each curve in the phase portrait represents a short trajectory, with its starting point and direction indicating the initial conditions and the direction of evolution over time. A closer look reveals significant differences in dynamics within region $u_2$ across the ODEs. While the curves in region $u_1 = [-2, 0] \times [-2, 0]$ consistently move from the bottom right to the top left in all phase portraits, the trajectories in region $u_2 = [2, 4] \times [-1, 1]$ exhibit drastically different behaviors. This indicates that trajectories originating from region $u_2$ are more divergent and thus more informative.

\noindent\textbf{Main Procedure.}
The proposed \method, illustrated in Figure~\ref{fig:pipeline}, comprises two key components: (1) Deep Sequential Decoder. This module predicts candidate ODEs by sampling sequences of grammar rules defined for symbolic ODE representation.
(2) Data Sampling Module. Using the proposed phase portrait sketching, this module selects a batch of informative ground-truth data points.

Throughout the training process, the reward for the predicted ODEs is computed using the queried data, and the decoder parameters are updated via policy gradient estimation. Among all sampled candidates, \method selects the ODE with the smallest loss value (as defined in Equation~\ref{eq:obj}) as the final prediction.

\noindent\textbf{Connection to Existing Approaches.}  Like~\citet{2023odeformer}, \method employs a Transformer-based decoder. However, unlike~\citet{2023odeformer}, which learns from fixed data, \method actively queries new data.
The learning objective of \method is inspired by~\citet{DBLP:conf/iclr/PetersenLMSKK21}, where both approaches guide the search for the optimal equation as a decision-making process over a sequence of tokens. 

Existing active learning methods, particularly in symbolic regression, have largely overlooked the chaotic behaviors inherent in dynamical systems.
For instance, \citet{DBLP:conf/aaai/JinH0HNDGC23} proposed a separate generative model for sampling informative data, assuming that input data within a small region should exhibit minimal output divergence. However, this assumption fails to hold in the context of dynamical systems.  Additionally, \citet{haut2023active} formulated an optimization problem based on the Query-By-Committee (QBC) method in active learning, to find those informative initial conditions. But the optimization needs to maintain a large set of data points, to account for the chaotic behaviors.  The rest of the discussion is provided in the Related Work.

\subsection{The Learning Pipeline}

\noindent\textbf{Data Assumption.}  Our method relies on the assumption that we can query a data oracle $\mathcal{O}$ by specifying the initial conditions $\mathbf{x}_0$ and discrete times $T=(t_1,\ldots,t_k)$. The oracle executes $\mathcal{O}(\mathbf{x}_0, T)$ and returns a (noisy) observation of the trajectory at the specified discrete times $T$. In science, this data query process is achieved by conducting real-world experiments with specified configurations. Recent work~\citep{chen2022generalisation,keren2023computational,DBLP:conf/gecco/HautPB23} also highlight the importance of having the oracle that can actively query data points, rather than learning from a fixed dataset.

\noindent\textbf{Expression Representation.}  
To enable the sequential decoder to predict an ODE by generating a sequence step-by-step, we extend the context-free grammar to represent an ODE as a sequence of grammar rules~\cite{DBLP:conf/icml/TodorovskiD97, DBLP:conf/dis/GecOBDT22, DBLP:conf/iclr/Sun0W023}. The grammar is defined by the tuple $\langle V, \Sigma, R, S\rangle$, where $V$ is a set of non-terminal symbols, $\Sigma$ is a set of terminal symbols, $R$ is a set of production rules and $S \in V$ is the start symbol.

More specifically, each component of the grammar is:  1) For the non-terminal symbols, we use $A$ to represent a sub-expression for $\frac{\dd x_1}{\dd t}$ and $B$ to represent a sub-expression for $\frac{\dd x_2}/{\dd t}$. For dynamical systems with $n$ variables, we use $n$ distinct non-terminal symbols. 2) The terminal symbols include the input variables and constants $\{x_1, \ldots, x_n, \mathtt{const}\}$. 3) The production rules correspond to mathematical operations. For example, the addition operation is represented as $A \to (A + A)$, where the rule replaces the left-hand symbol with the right-hand side. 4) The start symbol is redefined as ``$\phi \to A, B$'', where the comma notation indicates that $A$ and $B$ represent two separate equations in a two-variable dynamical system. Similarly, there will be $n$ non-terminal symbols connected by  $n-1$ comma 
for $n$-dimensional dynamical system.

Starting from the start symbol, different symbolic ODEs are constructed by applying the grammar rules in various sequences. An ODE is valid if it only consists of terminal symbols; otherwise, it is invalid. Figure~\ref{fig:pipeline}(b) provides an example of constructing the ODE $\frac{\dd x_1}{\dd t} = x_2$, $\frac{\dd x_2}{\dd t} = -0.9 \sin(x_1)$ from the start symbol $\phi \to A, B$ using a sequence of grammar rules. The replaced parts are color highlighted. Initially, the multiplication rule $B \to B \times B$ is applied, replacing the symbol $B$ in $f_2 = B$ with $B \times B$, resulting in $\phi \to A, B \times B$. Next, the rule $A \to x_2$ is applied, yielding $\phi \to x_2, B \times B$. Iteratively applying the rules, we eventually obtain $\phi \to x_2, c_1 \times \sin(x_1)$, which corresponds to one candidate ODE $\phi = (x_2, c_1 \sin(x_1))$. The coefficient $c_1=-0.9$ is obtained when fitting to the trajectory data. The procedure of coefficient fitting is described in Appendix~C ``Implementation of \method'' section.

\noindent\textbf{Sampling ODEs from Decoder.} 
The proposed \method is built on top of a sequential decoder, which generates different ODEs as a sequential decision-making process. The decoder can be RNN, LSTM, or the decoder-only Transformer. The input and output vocabulary is the set of allowed rules covering input variables, coefficients, and mathematical operators. 
Predicting ODEs involves using the decoder to sample a sequence of grammar rules, where each sequence corresponds to a candidate ODE using previously defined grammar. The objective of \method is to maximize the probability of sampling those ODEs that fit the data well. This is achieved through the REINFORCE objective, where the objective computes the expected reward of ODE to the data. In our formulation, the reward is evaluated on selected data by the phase portrait sketching module.

As shown in Figure~\ref{fig:pipeline}(a), the decoder receives the start symbol $s_0=$ ``$\phi\to A,B$'' and outputs a categorical distribution $p_{\theta}(s_{1}|s_0)$ over rules in the output vocabulary. This distribution represents the probabilities of possible next rules in the partially completed expression. One token is drawn from this distribution, $s_{1} \sim p(s_{1}|s_0)$, which serves as the prediction for the second rule and is used as the input for the next step. At $t$-th step, the predicted output from the previous step $s_t$ is used as the input for the current step. The decoder draws rule $s_{t+1}$ from the probability distribution $s_{t+1}\sim p_{\theta}(s_{t+1}|s_0,\ldots, s_t)$. This process iterates until maximum steps are reached, with a probability of $p_{\theta}(s)=\prod_{i=1}^{m-1} p_{\theta}(s_{i}|s_{1},\ldots, s_{i-1})$. The sampled sequence is converted into an expression following the definition previously described in ``Expression Representation''.

\noindent\textbf{Active Query Data with Phase Portrait Sketching.} To evaluate the goodness-of-fit of generated ODEs from the decoder and differentiate which one is better, we propose comparing the phase portrait of predicted ODEs. We sketch the phase portrait using collections of short trajectories, all starting from the same initial conditions and sharing the same time sequence.

Following our discussion in the ``Motivation'' section, a region is considered informative for distinguishing between two candidate ODEs if their sketched phase portraits differ. To identify such regions, we randomly sample several and sketch the phase portraits for all candidate ODEs within each. The most informative region is then selected, and we query the data oracle (noted as $\mathcal{O}$) for the ground-truth trajectory in that region.

Formally, assume we are given $M$ ODEs, $\{\phi_1,\ldots, \phi_M\}$, and $K$ randomly selected regions, $\{u_1, \ldots, u_K\}$. Each region $u_k$ is a Cartesian product of $n$ intervals, expressed as $u_k = [a_1, b_1] \times \cdots \times [a_n, b_n]$. To sketch the dynamics of candidates in the region $u_k$, we uniformly sample $L$ points in $u_k$, $\mathbf{x}^1, \ldots, \mathbf{x}^L$,  as initial conditions. For region $u_k$, the trajectory $\tau_{m,k,l} = (\mathbf{x}^l, \hat{\mathbf{x}}(t_1), \ldots, \hat{\mathbf{x}}(t_k))$ is generated by the expression $\phi_m$, starting from the $l$-th initial condition $\mathbf{x}^l$ and evolving over time according to the numerical integration $\hat{\mathbf{x}}(t_i) = \mathbf{x}^l + \int_0^{t_i} \phi_m(\mathbf{x}(t), \mathbf{c}) \, \mathrm{d}t$ for $t_i \in \{t_1, \ldots, t_k\}$. The resulting $L$ short trajectories form a sketched phase portrait for ODE $\phi_m$ in the region $u_k$.

Two expressions, $\phi_m$ and $\phi_{m'}$, have similar sketches in region $u_k$ if their corresponding trajectories, starting from the same initial condition, are close. Specifically, this occurs when $\sum_{l=1}^L \|\tau_{m,k,l} - \tau_{m',k,l}\| \approx 0$. We define the pairwise informative score between $\phi_m$ and $\phi_{m'}$ in region $u_k$ as:
\begin{equation} \label{eq:if-score}
\mathtt{IF}(\phi_m, \phi_{m'}, u_k) = \frac{1}{L} \sum\limits_{l=1}^L \|\tau_{m,k,l} - \tau_{m',k,l}\|_2^2 
\end{equation}
The total informative score for a region (denoted as $\mathtt{IF}(u_k)$)  is the sum of the pairwise informative scores for every pair of candidate ODEs. The informative score for region $u_k$ is:
\begin{equation}
\mathtt{IF}(u_k) = \frac{1}{M} \sum_{m=1}^M \sum_{m' = m+1}^M \mathtt{IF}(\phi_m, \phi_{m'}, u_k)
\end{equation}
We select the region with the highest informative score, denoted $u^* \gets \arg\max_{k=1}^K \mathtt{IF}(u_k)$. A batch of $m$ initial conditions, $\{\mathbf{x}_1, \ldots, \mathbf{x}_m\}$, is then sampled from region $u^*$, and the data oracle $\mathcal{O}(\mathbf{x}_i, T)$ is queried with the given initial conditions. The obtained ground-truth trajectories are used to compute the reward function for the objective, which in turn updates the model's parameters. In practice, the relative size of the regions and the number of sampled regions are set as hyper-parameters in the experiments.

\noindent\textbf{Policy Gradient-based Training.} The REINFORCE objective that maximizes the expected reward is
\begin{equation*}
J(\theta):=\mathbb{E}_{s \sim p_{\theta}(s)}[\mathtt{reward}(s)]
\end{equation*}
where $p_\theta(s)$ is the probability of sampling sequence $s$ and $\theta$ represents the parameters of the decoder. Following the REINFORCE policy gradient algorithm~\cite{DBLP:journals/ml/Williams92}, the gradient \textit{w.r.t.} the objective $\nabla_{\theta}J(\theta)$ is estimated by the empirical average over the samples from the probability distribution $p_\theta(s)$. We first sample $N$ sequences $(s^1,\ldots, s^N)$, and an unbiased estimation of the gradient of the objective is computed as: 
\begin{equation*}
\nabla_{\theta}J(\theta) \approx \frac{1}{N}\sum_{i=1}^N \mathtt{reward}({s}^i) \nabla_{\theta}\log p_{\theta}({s}^i)
\end{equation*}
The parameters of the decoder are updated using the estimated policy gradient value. This update process increases the probability of generating high goodness-of-fit ODEs. Detailed derivations are presented in Appendix C.

%% file: tex/4.related.tex
\section{Related Work} \label{sec:related}

\textbf{AI-driven Scientific Discovery.} Artificial intelligence has increasingly been employed to accelerate discoveries in learning ordinary and partial differential equations directly from data~\citep{brunton2016sparse,PhysRevE.100.033311,doi:10.1098/rspa.2018.0305,iten2020discovering,DBLP:conf/nips/CranmerSBXCSH20,Raissi20Fluid,raissi2019physics,Liu21AIPoincare,nanovoid_tracking,chen2018neural}. 

\begin{table*}[!t]
     \centering
     \begin{tabular}{r|lccc|cccc}
    \hline
        &    \multicolumn{4}{c|}{Strogatz dataset ($\sigma^2=0,\alpha=0$)}   & \multicolumn{4}{c}{ODEbase dataset ($\sigma^2=0,\alpha=0$)}  \\
        & $n=1$ & $n=2$ & $n=3$ & $n=4$  & $n=2$ & $n=3$ & $n=4$ &  $n=5$\\
         \hline
            SPL &  $0.787$ & $0.892$ & $1.921$ & $2.865$   &0.867 & 2.17&  4.75& 13.16\\
            E2ETransformer & $6.47E{-}4$ & $1.620$ & $T.O.$ & $T.O.$ & 0.757 &$T.O.$ & $T.O.$ & $T.O.$\\
            ProGED & $0.129$ & $0.666$  & $2.68$ &  $3.856$ & 0.317 &2.134 & $T.O.$ & $T.O.$\\
            SINDy & $1.90E{-}4$ & $\textbf{0.217}$ & $1.539$ & $4.810$ & 0.521 & 2.112 & $8.334$ & 52.12 \\
            ODEFormer &  $0.0303$  &  $0.9261$  &  $1.033$ & $1.010$ & 0.213 & 0.245 & $1.213$ &  $3.148$\\ 
           \method(ours)&  $\mathbf{2.06E{-}6}$   & $0.2912$  & $\textbf{1.011}$ & $\textbf{0.521}$ & $\mathbf{0.1318}$  & $\mathbf{0.1306}$  &$\mathbf{1.046}$ & $\mathbf{3.054}$\\
    \hline
    \end{tabular}
    \caption{On the \textit{noiseless} datasets with regular time sequence ($\sigma^2=0,\alpha=0$), Median NMSE is reported over the best-predicted expression found by all the algorithms. Our \method method can discover the governing expressions with smaller NMSE values than baselines, under the noiseless setting. T.O. means termination with a 24-hour limit.}
     \label{tab:nmse}
 \end{table*}

\noindent\textbf{Symbolic Regression for ODEs.} Symbolic regression, traditionally used to identify algebraic equations between input variables and output labels, has been extended to discover ODEs. A key ingredient is gradient matching, which approximates labels for symbolic regression by using finite differences of consecutive states along a trajectory~\citep{DBLP:conf/iclr/Sun0W023,DBLP:journals/kbs/BrenceTD21,DBLP:conf/iclr/QianKS22,DBLP:conf/dis/GecOBDT22}. Recent methods, such as SINDy and its extensions~\citep{brunton2016sparse,egan2024automatically}, leverage sparse regression techniques to directly learn the structure of ODEs and PDEs from data. They perform particularly well with trajectory data sampled at small, regular time intervals, where the approximations closely align with true derivatives.

\noindent\textbf{Neural Networks Learns Implicit ODEs.} This research direction involves learning ODE implicitly. Early work employed Gaussian Processes to model ODEs~\citep{DBLP:conf/icml/HeinonenYMIL18}. Neural ODEs further advanced the field by parameterizing ODEs with neural networks, enabling training through backpropagation via ODE solvers~\citep{chen2018neural}. Physics-informed neural networks integrate physical knowledge, such as conservation laws, into the modeling process~\citep{raissi2019physics}. Meanwhile, Fourier neural operators use neural networks to learn the functional representation~\citep{DBLP:conf/iclr/LiKALBSA21}.

\noindent\textbf{Active Learning} aims to query informative unlabeled data to accelerate convergence with fewer samples~\citep{DBLP:conf/colt/WagenmakerJ20,DBLP:journals/jmlr/ManiaJR22,DBLP:conf/iclr/SenerS18,DBLP:conf/iclr/AshZK0A20}. In symbolic regression, query-by-committee strategies have been explored to actively query data for discovering algebraic equations~\citep{DBLP:conf/gecco/HautBP22,DBLP:conf/gecco/HautPB23}. For example, \citet{DBLP:conf/aaai/JinH0HNDGC23} proposed a method that learns uncertainty distributions using neural networks and queries data with high uncertainty. However, all these methods largely overlooked the chaotic behaviors inherent in dynamical systems.

%% file: tex/5.exp.tex
\section{Experiments}
This section shows our \method can find ODEs with the smallest errors (Normalized MSE) among all competing approaches, under noiseless, noisy, and irregular time settings (see Table~\ref{tab:nmse} and Table~\ref{tab:nmse-extend}).
Compared to the baselines, our \method data query strategy requires fewer data and attains a better ranking of the TopK candidate ODEs (see Table~\ref{tab:diff-active}). 

\subsection{Experimental Settings}
\noindent\textbf{Datasets.} We consider 2 datasets of multivariate variables, including  (1) Strogatz dataset~\cite{2023odeformer} of 80 instances, collected from the Strogatz textbook~\cite{strogatz2018nonlinear}. It is formalized as a benchmark dataset by~\cite{2023odeformer}.  (2) ODEBase dataset~\cite{luders2022odebase} of 114 instances, containing equations from chemistry and biology. Each dataset is further partitioned by the number of variables contained in the ODE.

We consider 3 different conditions: (1) regular time noiseless condition, (2) regular time noisy condition, and (3) irregular time condition. In the noiseless setting, the obtained data is exactly the evaluation of the ground-truth expression.
In the noisy setting, the obtained data is further perturbed by Gaussian noise. We add multiplicative noise by replacing each $\mathbf{x}(t_i)$ with $(1+\varepsilon) \mathbf{x}(t_i)$, and $\varepsilon$ is sampled from a zero mean multivariate Gaussian distribution with diagonal variances $\mathtt{diag}(\sigma^2,\ldots,\sigma^2)$. The noise rate is determined by $\sigma^2$. For both noiseless and noisy settings, the data points are sampled at regular time intervals. In the irregular time setting, we first generate the regular time sequence and drop a fraction with probability $\alpha$. The rate of time irregularity is determined by $\alpha$.  

\begin{table*}
    \centering
    \begin{tabular}{r|cccc|cccc}

\hline
        &    \multicolumn{4}{c|}{{Noisy} Strogatz datasets ($\sigma^2=0.01,\alpha=0$)}   & \multicolumn{4}{c}{Irregular Strogatz dataset ($\sigma^2=0,\alpha=0.1$)}   \\
        & $n=1$ & $n=2$ & $n=3$ & $n=4$  &   $n=1$ & $n=2$ & $n=3$ & $n=4$ \\ \hline
             SPL& $0.938$ & $1.019$  & $2.915$  & $3.068$ &  $0.127$ &   $0.526$  & $3.196$ & $4.193$  \\
         SINDy & $6.4E{-}3$ & 4.152  & $2.498$ & 5.21 & $6.66E{-}4$ & 0.472  & $0.827$  & $4.163$\\
         ProGED &  $0.121$  & $0.658$ & $3.673$ & $3.856$  & $0.134$ & $0.769$ & $ 2.766$ & $4.181$\\
     
             ODEFormer &  $0.139$  &  $0.621$  & $2.392$ & $0.812$ &$0.031$  &  $1.036$  & 1.51 & 1.011  \\
         \method(ours)&  $\textbf{7.75E-4}$   & $\textbf{0.369}$  &  $\textbf{1.381}$ & $\mathbf{0.657}$ &  $\textbf{1.06E{-}6}$   & $\textbf{0.215}$  & $\textbf{ 1.012}$ & $\textbf{0.947}$ \\
    \hline
    \end{tabular}
    \caption{On the Strogatz dataset, the Median NMSE is reported over the best-predicted expression found by all the algorithms under noisy or irregular time sequence settings.}
    \label{tab:nmse-extend}
\end{table*}

\noindent\textbf{Baselines.} We consider a line of recent works for symbolic equation discovery as our baselines. The methods using passive data query strategy are as follows: (1) SINDy~\cite{brunton2016sparse}, (2) ODEFormer~\cite{2023odeformer}, (3)  Symbolic Physics Learner (SPL)~\cite{DBLP:conf/iclr/Sun0W023}, (4) Probabilistic grammar for equation discovery (ProGED)~\cite{DBLP:conf/dis/GecOBDT22}, 
(5) end-to-end Transformer (E2ETransformer)~\cite{DBLP:conf/nips/KamiennydLC22}.

\noindent\textbf{Evaluation.} For evaluating all the methods, we considered 3 different metrics: (1) goodness-of-fit using NMSE, (2) empirical running time of data querying step, and (3) ranking-based distance. The goodness-of-fit using the NMSE indicates how well the learning algorithms perform in discovering symbolic expressions. Given the best-predicted expression by each algorithm, we evaluate the goodness-of-fit on a larger testing set with longer time steps and a larger batch size of data. The median (50\%) of the NMSE is reported in the benchmark table. The full quantiles ($25\%, 50\%, 75\%$) of the NMSE are further provided.
The remaining details of the experiment settings are in Appendix D.

\subsection{Experimental Analysis}

\noindent\textbf{Goodness-of-fit Benchmark.} We summarize our \method on several challenging multivariate datasets with noiseless data in Table~\ref{tab:nmse}. It shows our \method attains the smallest median NMSE values on all datasets, against a line of current popular baselines. The performance of SPL and E2Etransformer drops greatly on irregular time sequences because the approximated time derivative becomes inaccurate when missing the intermediate sequence. Our \method does not suffer from that because it outputs the predicted trajectory and does not need to approximate the time derivative. Another reason is the decoder with massive parameters can better adapt to actively collected datasets.

 \begin{figure}[!t]
    \centering
    \includegraphics[width=0.62\linewidth]{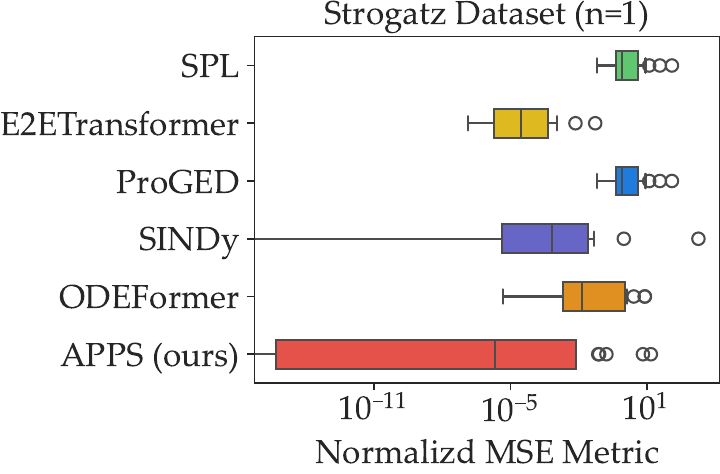}
    \includegraphics[width=0.32\linewidth]{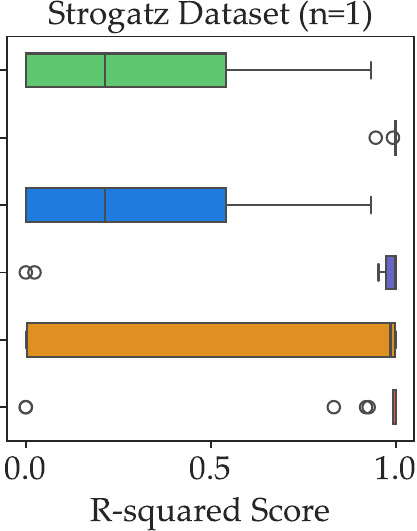}
    \caption{On the selected data (Strogatz dataset with $n=1$), quartiles of  NMSE and $R^2$ scores of the learning algorithms.}
    \label{fig:quartile-strogatz-n1}
\end{figure} 
\noindent\textbf{Noisy and Irregular Time Settings.} We examine the performance of predicting trajectories in the presence of noise and irregular time sequences in Table~\ref{tab:nmse-extend}. The ground-truth trajectory is subject to Gaussian noise with zero mean and $\sigma^2=0.05$, and an irregularly sampled sequence where $50\%$ of evenly spaced points are uniformly dropped. The predicted trajectory by each algorithm is compared against the ground truth, utilizing identical initial conditions. Our \method still attains a relatively smaller NMSE against baselines under the two settings. 

\noindent\textbf{Quantiles of Evaluation Metrics.}
We further report the quantiles of the NMSE metric in Figure~\ref{fig:quartile-strogatz-n1} to assist the result in Table~\ref{tab:nmse}(a). Note that we cut off the negative values as zero when demonstrating $R^2$ score. The two box plots in Figure~\ref{fig:quartile-strogatz-n1} show the proposed \method is consistently better than the baselines in terms of the full quantiles ($25\%, 50\%, 75\%$) of the NMSE metric.

\noindent\textbf{Benchmark with other Active Strategies.} 
Two baseline methods using active learning strategy are: (1)  query-by-committee (QbC) proposed in~\cite{DBLP:conf/gecco/HautBP22,DBLP:conf/gecco/HautPB23}.  (2)  Core-Set~\cite{DBLP:conf/iclr/SenerS18} proposes to sample diverse data. These methods were originally proposed with different neural networks, thus we evaluate these different active learning methods using the same decoder in our \method. Current active learning methods are not directly available for evaluation in our problem setting (in Equation~\ref{eq:obj}), so we re-implement these query strategies with the new problem setting. 

The running time of the data querying step measures the efficiency of every active learning algorithm for this task. The ranking-based distance indicates if the ranking of many candidate expressions is exactly the same as evaluated on full data. If the predicted ODEs are ranked in the same order as the full data, then the ranking-based distance (Kendall tau score) will be close to zero.

In Table~\ref{tab:diff-active}, given a set of 20 predicted ODEs, we compare the TopK ranking (i.e., top 3) of predicted ODEs by each active learning strategy is the same as using full data.  We find both our phase portrait and QbC rank those predicted ODEs in proper ranking order. Our \method takes the least memory to locate the most informative region and is also time efficient because we only pick one region among all the available regions.
The QbC takes much more time because it finds every initial condition as an optimization problem over the input variables, which is solved by a separate gradient-based optimizer. 
CoreSet first runs a clustering algorithm over the ground-truth data and then samples a diverse set of initial conditions from each cluster. So the memory usage of Coreset is mainly determined by the first clustering step.

 \begin{table}[!t]
     \centering
     \begin{tabular}{r|ccr}
    \hline
        &Ranking-based  &  Running  & Peak  \\  
        & distance ($\downarrow$) &  Time ($\downarrow$) & Memory ($\downarrow$) \\ \midrule
         \method (ours)&  $\mathbf{0.08}$ & $\mathbf{5.2}$ sec & $\mathbf{3.76}$ MB \\
         QbC & $0.13$ &  $13.4$ sec & $51$ MB \\
         CoreSet &  $0.22$ & $4.3$ sec & $2.74$ GB \\
         \hline
    \end{tabular}
   \caption{Ranking comparison with different active learning strategies. \method shows a smaller ranking-based distance than other strategies, which is better for ranking those best-predicted expressions. Also \method takes less memory consumption and less computational time because the sketching step itself is lightweight.}
    \label{tab:diff-active}
\end{table}

%% file: tex/6.conclude.tex
\section{Conclusion}
In this paper, we introduced \method, a novel approach for discovering ODEs from trajectory data. By actively reasoning about the most informative regions within the phase portrait of candidate ODEs, \method overcomes the limitations of passively learned methods that rely on pre-collected datasets. Our approach also reduces the need for extensive data collection while still yielding highly accurate and generalizable ODE models.
The experimental results demonstrate that \method consistently outperforms baseline methods, achieving the lowest median NMSE across various datasets under both noiseless and noisy conditions.

%% file: tex/10.3.phase.tex
\section{Extended Preliminaries} \label{apx:phase}

\textbf{Phase Plane.} The phase plane is a visual display of solutions of differential equations. 
Given an ODE, its solutions are a family of functions, which can be graphically plotted in the phase plane. At point $(x_1,x_2)$, we draw a vector representing the derivatives of the point with respect to the time variable, that is $(\dd x_1/\dd t, \dd x_2/\dd t)$. With sufficient of these arrows in place the system behavior over the regions of the place can be visualized and the long-term behavior can be quantitatively determined, like the limit cycles and attractors. The obtained entire figure is known as the \textit{phase portrait}. It is a geometric representation of all possible trajectories from the corresponding ODE. One can interpret the phase plane in terms of dynamics. The solution of ODE corresponds to a trajectory of a point moving on the phase plane with velocity.

\noindent\textbf{Butterfly Effect.}  In the context of ordinary differential equations (ODEs), the butterfly effect implies the \textit{sensitive dependence on initial conditions} in dynamical systems. It means that small differences in the initial state of a system can lead to vastly different trajectories over time.

Mathematically, let $\mathbf{x}_0$ and $\mathbf{x}'_0$ be two initial conditions that are very close to each other, i.e., $\|\mathbf{x}_0 - \mathbf{x}_0'\| \le \delta $. The butterfly effect refers to the situation where the distance between the corresponding trajectories, $\mathbf{x}(t)$ and $\mathbf{x}'(t)$, grows exponentially over time:
\begin{equation*}
\|\mathbf{x}(t) - \mathbf{x}'(t)\| \approx \exp({\lambda t})\|\mathbf{x}_0 - \mathbf{x}_0'\|,
\end{equation*}
where $\lambda $ quantifies the rate at which two nearby trajectories diverge. If $\lambda $ is positive, small initial differences grow exponentially over time, implying that even a tiny perturbation (like a butterfly flapping its wings) can lead to dramatically different outcomes in a chaotic system.

In this research, where the task is to query informative data to rank a given list of ODEs,  this phenomenon implies the drawn data (i.e., initial conditions) are less informative but its close neighbor is highly informative. To avoid this, we can consider evaluating more initial conditions, which demand huge space usage and slow time computation.

\noindent\textbf{Different trajectories never intersect in the phase plane.}
Solutions of ODEs are uniquely defined by initial conditions except at some special points in the phase plane. Trajectories in the phase plane cannot cross (except at some special points) as this would be equivalent to non-uniqueness of solutions. The special points are fixed points or singular points where trajectories start or end.

The existence and uniqueness theorem has an important corollary: different trajectories never intersect. If two trajectories did intersect, then there would be two solutions starting from the same point (the crossing point), and this would violate the uniqueness part of the theorem. In other words, a trajectory cannot move in two directions at once. Because trajectories cannot intersect, phase portraits always have a well-groomed look to them.

\begin{theorem-no}[Existence and Uniqueness~\cite{coddington1955theory}]
Consider the initial value problem $\dot{\mathbf{x}} = \mathbf{f}(\mathbf{x})$, $\mathbf{x}(0) = \mathbf{x}_0$.
Suppose that $ \mathbf{f} $ is continuous and that all its partial derivatives 
$ {\partial {f}_i}/{\partial x_i} $, $ i, j = 1, \ldots, n $, are continuous for $ \mathbf{x} $ in some open connected set $ D \subset \mathbb{R}^n $. Then for $ \mathbf{x}_0 \in D $, the initial value problem has a solution $ \mathbf{x}(t) $ on some time interval $ (a,b) $ about $ t = 0 $, and the solution is unique.
\end{theorem-no}

\noindent\textbf{Active Learning}  is a machine learning method in which the learning algorithm inspects unlabeled data and interactively chooses the most informative data points to learn. The goal of active learning algorithms is to learn an accurate predictor with as little total data as possible. There are two
standard settings: pool-based and streaming-based settings. In the pool-based setting, the learner is
provided with a pool of unlabeled data, from which the interactively selects the most informative points and asks for their label. In the streaming-based setting,
the learner receives a sequence of unlabeled points and decides on whether to request the label of the current point. In this research, we only consider pool-based active learning algorithms.

According to the active learning~\cite{medina2023active,DBLP:conf/l4dc/Buisson-FenetST20,DBLP:conf/gecco/HautBP22,DBLP:conf/aaai/JinH0HNDGC23}, the input is the initial condition $\mathbf{x}_0\in\mathbb{R}^n$ and the output is the obtained trajectory $(\mathbf{x}_{t_1},\ldots,\mathbf{x}_{t_k})\in\mathbb{R}^{n\times k}$. In the formulation of the Query-by-Committee method, they define the uncertainty at an input point as the variance of the predictions of the committee members.


%% file: tex/10.1.extend-method.tex

%% file: tex/10.2.implement.tex
\section{Extended Explanation of \method method} \label{apx:implement}





\paragraph{Data-availability Assumption.} 
A crucial assumption behind the success of \method is the availability of a  Data Oracle $\mathcal{O}$ that returns a (noisy) observation of the trajectory with a specified initial condition and a sequence of discrete times.
Such a data oracle represents conducting controlled experiments in the real world, which can be expensive. 
This differs from the current symbolic regression, where a dataset is obtained prior to learning. 

\paragraph{Vocabulary Construction.} Given the set of math operators and variables $O_m=\{+,-,\times,\div,\sin\ldots\}\cup\{x_1,\ldots, x_n,\mathtt{const}\}$, where $\mathtt{const}$ indidates the coefficients in the expressions. Following the definition of grammar in Section~\ref{sec:method}. For example, given $O_m=\{+,-,\times,
\div \}\cup\{x_1, x_2,\mathtt{const}\}$, we construct the following grammar rules:
\begin{verbatim}
A -> (A + A)
A -> (A - A)
A -> A * A
A -> A / A
A -> x1
A -> x2
A -> const
B -> (B + B)
B -> (B - B)
B -> B * B
B -> B / B
B -> x1
B -> x2
B -> const
\end{verbatim}
The non-terminal symbol ``$A$''  denotes a subexpression in $\mathit{d} x_1$ and similarly ``$B$''  denotes a subexpression in $\mathit{d} x_2$.
Each of them will be a unique token of the input vocabulary of the decoder and will be mapped to a distinct vector in the first embedding layer of the decoder. The above rules also form the output vocabulary of the decoder, where the neural decoder predicts a categorical distribution over the list of grammar rules.

The discovery path is to build algorithms that mimic human scientific discovery, which has achieved tremendous success in early works~\cite{DBLP:conf/ijcai/Langley77,DBLP:conf/ijcai/Langley79,DBLP:conf/ijcai/LangleyBS81}. Recent work~\cite{chen2022generalisation,keren2023computational,DBLP:conf/gecco/HautBP22,DBLP:conf/gecco/HautPB23} also pointed out the importance of having a data oracle that can actively query data points, rather than learning from a fixed dataset.

\paragraph{Sequential Decision Making Formulation.} 
We consider an undiscounted MDP with a finite horizon. 
The state space $\mathcal{S}$ is the set of all possible sequences of rules with maximum steps.
The action space $\mathcal{A}$ is the set of grammar rules.
The $t$-th step state $s_t$ is the sequence of sampled rules before the current step $t$, i.e., $s_t:=(s_1,\ldots,s_t)$. 
The action $a_t$ is the sampled single rule, $a_t:=s_{t+1}$.

The loss function of \method is informed by the REINFORCE algorithm~\cite{DBLP:journals/ml/Williams92}, which is based on the log-derivative property: 
\begin{align*}
    \nabla_{\theta}p_{\theta}(s) = p_{\theta}(s) \nabla_{\theta}\log p_{\theta}(s)
\end{align*}
 where $p_{\theta}(s)\in(0,1)$ represents a probability distribution over input $s$ with parameters $\theta$ and notation $\nabla_\theta$ is the partial derivative with respect to $\theta$.
In our formulation, let $p_{\theta}(s)$ denote the probability of sampling a sequence of grammar rules $s$ and $\mathtt{reward}(s)=1/(1+\mathtt{NMSE}(\phi))$. Here $\phi$ is the corresponding expression constructed from the rules $s$ following the procedure in Section~\ref{sec:method}. The probability $p_{\theta}(s)$ is modeled by the decoder modules.
The objective described in Equation~\ref{eq:obj} is translated to maximize the expected reward of the sampled sequences from the decoder:
\begin{equation*}
    \arg\max_{\theta}\;\mathbb{E}_{s\sim p_\theta(s)}[\mathtt{reward}(s)]  
\end{equation*}
Based on the REINFORCE algorithm, the gradient of the objective can be expanded as:
\begin{equation*}
\begin{aligned}
\nabla_{\theta}\mathbb E_{s\sim p_{\theta}(s)}&[\mathtt{reward}(s)] = \nabla_{\theta}\sum_{s\in \Sigma}{\mathtt{reward}(s)p_{\theta}(s) }\\
		& = \sum_{s\in \Sigma}{\mathtt{reward}(s)\nabla_{\theta}p_{\theta}(s) }\\
        & = \sum_{s\in \Sigma}{\mathtt{reward}(s) p_{\theta}(s)} \frac{\nabla_{\theta}p_{\theta}(s)}{p_{\theta}(s)}\\
		& = \sum_{s\in \Sigma}{\mathtt{reward}(s)p_{\theta}(s) \nabla_{\theta}\log p_{\theta}(s)}\\
		& = \mathbb E_{s\sim p_{\theta}(s) }\left[\mathtt{reward}(s)\nabla_{\theta}\log p_{\theta}(s) \right]\\
\end{aligned}
\end{equation*}
where $\Sigma$ represents all possible sequences of grammar rules sampled from the decoder.
The above expectation can be estimated by computing the averaged over samples drawn from the distribution $p_{\theta}(s)$.
 We first sample several times from the decoder module and obtain $N$ sequences $(s^1,\ldots, s^N)$, an unbiased estimation of the gradient of the objective is computed as: $\nabla_{\theta}J(\theta)\approx \frac{1}{N}\sum_{i=1}^N\mathtt{reward}(s^i)\nabla_{\theta}\log p_{\theta}(s^i)$.
In practice, the above computation has a high variance. To reduce variance, it is
common to subtract a baseline function $b$ from the reward. In this study, we choose the baseline function as the average of the reward of the current sampled batch expressions. Thus we have:
 \begin{align*}
\nabla_{\theta}J(\theta)&\approx \frac{1}{N}\sum_{i=1}^N(\mathtt{reward}(s^i)-b)\nabla_{\theta}\log p_{\theta}(s^i), 
\end{align*}
where $b=\sum_{i=1}^N \mathtt{reward}(s^i)$. Based on the description of the execution pipeline of the proposed \method, we summarize every step in Algorithm~\ref{alg:main}.

\begin{algorithm*}[!t]
   \caption{Active Discovery of Ordinary Differential Equations via Phase Portrait Sketching. }\label{alg:main}  
   \begin{algorithmic}[1]
   \Require{Defined expression grammar; neural sequential decoder; data oracle for the ground-truth ODE $\mathcal{O}$; maximum learning epoch $\mathtt{\#epochs}$; discrete time steps $T=(t_1,\ldots, t_k)$.}
   \Ensure{The best-predicted ODE.}
   \State initialize the set of best predicted ODEs $\mathcal{Q}\gets \emptyset$; \Comment{initialization}
   \State  randomly draw data $D$ from Oracle $\mathcal{O}$.
     \For{$t \gets 1 \textit{ to } \mathtt{\#epochs} $}
     
   \State  sample $N$ sequences $\{s_1,\ldots, s_N\}$ from the sequential decoder. \Comment{in Figure~\ref{fig:pipeline}(a)}
   
   \State construct   $N$ ODEs $\{\phi_i\}_{i=1}^N$ for each sequence from defined grammar.  \Comment{in Figure~\ref{fig:pipeline}(b)}
   \State fit coefficients in each expression with data $c_i\gets\mathtt{Optimize}(\phi_i, D)$.   
    \State find a region $u$ of high uncertainty with phase portrait sketching. \Comment{in Figure~\ref{fig:pipeline}(c)}
   \State draws trajectory data from Oracle in the selected region $D_u\gets\mathcal{O}(u, T)$. 
   \State computes reward using data $D_u$.
   \State save all expressions into $\mathcal{Q}$; 
    \State save new data $D_u$ into $D$.
   \State applies policy gradient to the parameters of the decoder.
    
    \EndFor
  \State \Return   the expression in $\mathcal{Q}$ with best goodness-of-fit on data $D$.
\end{algorithmic}
\end{algorithm*}
 
\subsection{Implementation of \method}

In the experiments, we use an embedding layer, a multi-head self-attention layer, and finally softmax layer as the decoder. The dimension of the input embedding layer and the hidden vector is configured as $256$.  We use the Adam optimizer as the gradient descent algorithm with a learning rate of $0.009$. The learning epoch is configured as $50$. The maximum sequence of grammar rules is fixed to be $20$. The batch size of expressions sampled from the decoder is set as $100$.

When fitting the values of coefficients in each expression, we sample a batch of data with batch size $1024$ from the data Oracle.  The open constants in the expressions are fitted on the data using the BFGS optimizer\footnote{\url{https://docs.scipy.org/doc/scipy/reference/optimize.minimize-bfgs.html}}. 

We use a multi-processor Python library \texttt{pathos} to fit multiple expressions in parallel using $20$ CPU cores. This greatly reduced the total time of the coefficients fitting step. The rest of the implementation details are available in the code implementation.

In terms of numerical integration, we use the fourth-order Runge–Kutta (RK45) method to compute the trajectory data of the specified ODEs. Other numerical integration algorithms in \texttt{sicpy.integrate} are implemented with adaptive time step size, which makes it very slow when fitting the coefficients in the candidate ODEs.

Every candidate ODE is represented by the Sympy Lambdify function. We implement the Runge-Kutta function instead of using the \texttt{Scipy.integrate.solve\_ivp} function, where the latter has internally used the adaptive step size and is extremely slow for specific candidate ODEs. In our experiments, we find that the \texttt{Scipy.integrate.solve\_ivp} API cannot return a trajectory when running together with the coefficient fitting steps for more than 12 hours.

\begin{figure*}
\centering
    \includegraphics[width=0.99\linewidth]{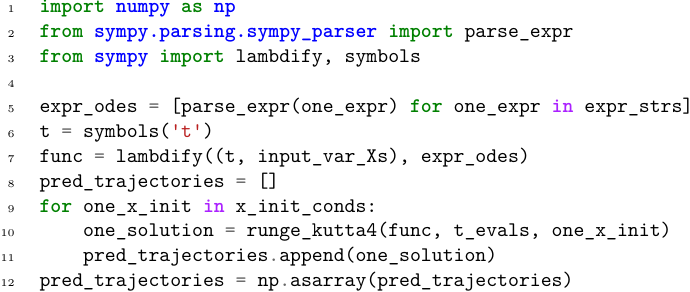}

\caption{Implemented 4th order Runge Kutter method.} \label{fig:runge-kutter}
\end{figure*}

In Figure~\ref{fig:runge-kutter}, given one predicted ODE \texttt{expr\_strs}, which is represented as an array of expressions of length $n$, the following lines of code compute the predicted trajectory \texttt{pred\_trajectories} for a batch of initial conditions \texttt{x\_init\_conds}. 
\texttt{input\_var\_Xs}  the list of symbols of variables: $[x_1,\ldots, x_n]$.
We can then compare if the predicted trajectories are close to the ground-truth trajectories using the NMSE metric, given the same set of initial conditions.

\paragraph{Hyper-parameter Configuration} An expression containing placeholder symbol $A$ or containing more than $20$ open constants is not evaluated on the data, the fitness score of it is $-\infty$.  In terms of the reward function in the policy gradient objective, we use $\mathtt{reward}(s)=\frac{1}{1+\mathtt{NMSE}(\phi)}$. The normalized mean-squared error metric is further defined in Equation~\ref{eq:loss-function}. We set the relative size of the interval of the region to be $1/4$, meaning the length of every edge of the region is $1/4$ of the original interval for each variable. We set the number of regions to be $10$.  The region is randomly generated by first generating the leftmost point in the original variable interval and then applying each edge of the region with a given relative length.

The deep network part is implemented using the most recent version of Pytorch, the expression evaluation is based on the Sympy library, and the step for fitting open constants in expression with the dataset uses the Scipy library. The visualization of the phase portrait is from \footnote{\url{https://phaseportrait.github.io/}} and the part of the experimental visualization is borrowed from \footnote{\url{https://github.com/sdascoli/odeformer/blob/main/ODEFormer_demo.ipynb}}.

\subsection{Limitation and Broader Impact} \label{apx:limit-and-impact} 

\noindent\textbf{Limitation} The proposed \method that generates phase portraits can suffer from resolution issues. Fine details of the dynamics might be missed if the region width is too large or the number of sampled points in each region is too small. It is also unclear if the proposed phase portrait sketching idea is applicable to partial differential equations.

\noindent\textbf{Broad Impact} The proposed \method can be useful for scientists to actively discover governing laws from data. It will accelerate the discovery process compared to passive learning algorithms.

%% file: tex/10.4.expset.tex
\section{Experiment Settings} \label{apx:exp-set}

\subsection{Baselines} \label{apx:exp-baseline}
For the baselines of the ODE discovery task, we consider
\begin{itemize}[leftmargin=*]
\item SINDy~\cite{brunton2016sparse}\footnote{\url{https://github.com/dynamicslab/pysindy}} is a popular method using a sparse regression algorithm to find the differential equations.
 \item ProGED~\cite{DBLP:journals/kbs/BrenceTD21}\footnote{\url{https://github.com/brencej/ProGED}} uses probabilistic context-free grammar to search for differential equations. ProGED first samples a list of candidate expressions from the defined probabilistic context-free grammar for symbolic expressions. Then ProGED fits the open constants in each expression using the given training dataset. The equation with the best fitness scores is returned.
\item ODEFormer~\cite{2023odeformer}\footnote{\url{https://github.com/sdascoli/odeformer}} is the most recent framework that uses the transformer for the discovery of ordinary differential equations. We use the provided pre-trained model to predict the governing expression with the dataset. We execute the model 10 times and pick the expression with the smallest NMSE error. The dataset size is $500$, which is the largest dataset configuration for the ODEFormer.
\end{itemize}

In principle, the ODE discovery task can be formulated as a symbolic regression task where the input is $\mathbf{x}_t$ and the output is directly $\dot{\mathbf{x}}_t$. Given the trajectory data $(\mathbf{x}_0,\mathbf{x}(t_1),\ldots, \mathbf{x}(t_n))$, its output label is approximated by computing: 
\begin{equation*}
\dot{\mathbf{x}}(t_i)\approx \frac{(\mathbf{x}(t_{i+1})-\mathbf{x}(t_{i}))}{(t_{i+1}-t_{i})}
\end{equation*}
In literature, this approach is called symbolic regression with gradient matching. We consider two representative baselines in the symbolic regression task:
\begin{itemize}[leftmargin=*]
\item Symbolic Physics Learner (SPL) is a heuristic search algorithm based on Monte Carlo Tree Search for finding optimal sequences of production rules using context-free grammars~\cite{DBLP:conf/icml/KamiennyLLV23,DBLP:conf/iclr/Sun0W023}\footnote{\url{https://github.com/isds-neu/SymbolicPhysicsLearner}}. 
It employs Monte Carlo simulations to explore the search space of all the production rules and determine the value of each node in the search tree. SPL consists of four steps in each iteration:   1) Selection. Starting at a root node, recursively select the optimal child (\textit{i.e.}, one of the production rules) until reaching an expandable node or a leaf node. 2) Expansion. If the expandable node is not the terminal, create one or more of its child nodes to expand the search tree. 3) Simulation. Run a simulation from the new node until achieving the result. 4) Backpropagation. Update the node sequence from the new node to the root node with the simulated result. To balance the selection of optimal child node(s) by exploiting known rewards (exploitation) or expanding a new node to explore potential rewards exploration, the upper confidence bound (UCB) is often used. 
    \item End to End Transformer for symbolic regression (E2ETransformer)~\cite{DBLP:conf/nips/KamiennydLC22}\footnote{\url{https://github.com/facebookresearch/symbolicregression}}. 
They propose to use a deep transformer to pre-train on a large set of randomly generated expressions. We load the shared pre-trained model. We provide the given dataset and the E2ETransformer infers 10 best expressions. We choose to report the expression with the best NMSE scores.
\end{itemize}

Note that the open link in NSODE~\cite{DBLP:conf/icml/BeckerKNPK23} has no code implementation by far. Thus, the NSODE approach is not included for comparison. Also, a recent method~\cite{DBLP:conf/aaai/JinH0HNDGC23} proposes a new learning framework for actively drawing data. However, their approaches are designed for searching algebraic equations, and the integration into differential equations is unclear. So their approach is also not considered for comparison in this work.

\begin{table*}[!t]
    \centering
    \begin{tabular}{r|cccc} 
    \toprule
           &\method & ProGED& ODEFormer  & SPL \\\midrule
         goodness-of-fit function & NegMSE &NegMSE &NegRMSE & NegRMSE  \\
         Training setting & \multicolumn{4}{l}{evolve for 1 sec with time step  $0.001$ sec and 100 random initial conditions} \\
         Testing setting & \multicolumn{4}{l}{evolve for 10 sec with time step $0.001$ sec and 100 random initial conditions} \\
      \#CPU for training & \multicolumn{4}{c}{20} \\
      \bottomrule
    \end{tabular}
    \caption{Major hyper-parameters settings for all the algorithms considered in the experiment.}
    \label{tab:baseline-hyper-config}
\end{table*}

\subsection{Evaluation Metrics} \label{sec:apx-evaluation}
The goodness-of-fit indicates how well the learning algorithms perform in discovering unknown ODEs.  The testing set contains new trajectories with new initial conditions, generated from the ground-truth ODE. we measure the goodness-of-fit of a predicted expression $\phi=[\phi_1,\ldots, \phi_m]$, by evaluating the mean-squared-error (MSE) and normalized-mean-squared-error (NMSE):
\begin{equation}\label{eq:loss-function}
\begin{aligned}
\text{NMSE}&=\frac{1}{\sigma^2}\frac{1}{n}\sum_{i=1}^n(\mathbf{x}(t_i)-\hat{\mathbf{x}}(t_i))^2,
\end{aligned}
\end{equation}
The empirical variance $\sigma=\sqrt{\frac{1}{n}\sum_{i=1}^n \left(\mathbf{x}_i-\frac{1}{n}\sum_{i=1}^n \mathbf{x}_i\right)^2}$. 
We use the NMSE as the main criterion for comparison in the experiments and present the results on the remaining metrics in the case studies. 
The main reason is that the NMSE is less impacted by the output range. The output ranges of expression are dramatically different from each other, making it difficult to present results uniformly if we use other metrics.

Prior work~\cite{DBLP:conf/iclr/PetersenLMSKK21} further proposed a coefficient of determination $R^2$-based score over a group of expressions in the dataset, as a statistical measure of whether the best-predicted expression is almost close to the ground-truth expression. An $R^2$ of 1 indicates that the regression predictions perfectly fit the data~\cite{nagelkerke1991note}. The $R^2$ score is computed as follows:
\begin{align}
R^2(\phi_i)=1-\texttt{NMSE}(\phi_i)
\end{align}

\subsection{Computational Resource} \label{apx:compute-set}
All the methods are running with Python 3.10 and on the same set of hardware where the CPU is Milan CPUs @ 2.45GHz, the RAM is set as 8GB, and the maximum running time is set as 24 hours. The extra necessary configuration is collected in the anonymous code repository. The hyper-parameter configurations of baselines are listed in Table~\ref{tab:baseline-hyper-config}.

\subsection{Extended Experimental Results} \label{apx:extra-exp}

The given set of best-predicted ODEs for Table~\ref{tab:diff-active} is shown in Figure~\ref{fig:eq-list}.
\begin{figure*}[!ht]
\begin{equation*}
\begin{aligned}
\phi_1&=(0.088x_1,     0.146\cos(x_1)) \\
\phi_{2}&=(-0.0107\sin(x_1),      0.712) \\
\phi_3&=( 2/\cos(x_1) + 0.1759\sin(x_1),    0.820) \\
\phi_4&=(0.645\sin(x_0) - 0.893\cos(x_0),    \\
& 0.2891\frac{x_1}{(x_1 - 0.153\cos(x_0))} + 0.213\frac{\sin(x_0)}{(x_1 - 0.153\cos(x_0))} - 0.0443\frac{\cos(x_0)}{(x_1 - 0.15\cos(x_0))}) \\
\phi_5&=(0.95\sin(x_0) - 0.29,    0.14153x_1 - 0.691\frac{\sin(x_1)}{\cos(x_0)} - 0.0673\cos(x_0) + 0.8825\frac{\cos(x_1)}{\cos(x_0)}) \\
\phi_6&=(1.63\sin(x_0),         x_0) \\
\phi_7&=(1.15\sin(x_0),         -70.56\cos(x_1)) \\
\phi_8&=(0.71,         -70.56\cos(x_1)) \\
\phi_9&=(0.99\cos(x_0),          -1.0x_0 - 39.53\sin(x_1) - 152.57\cos(x_1)) \\
\phi_{10}&=(-0.871\sin(x_1),         -3.826\sin(x_0) - 0.212\sin(x_1) + 4.311) \\
\phi_{11}&=(0.088 x_1,     0.146\cos(x_1)) \\
\phi_{12}&=(-0.010\sin(x_1),      0.712) \\
\phi_{13}&=(1.477e-5x_12/\cos(x_1) + 0.1759\sin(x_1),    0.8205) \\
\phi_{14}&=(0.645\sin(x_0) - 0.893\cos(x_0),    \frac{0.289x_1}{(x_1 - 0.15\cos(x_0))} + \frac{0.21\sin(x_0)}{(x_1 - 0.153\cos(x_0))} - \frac{0.044\cos(x_0)}{(x_1 - 0.1532\cos(x_0))}) \\
\phi_{15}&=(0.953\sin(x_0) - 0.2922,    0.1415x_1 - 0.691\sin(x_1)/\cos(x_0) - 0.0673\cos(x_0) + 0.882\frac{\cos(x_1)}{\cos(x_0)}) \\
\phi_{16}&=(1.631\sin(x_0),         x_0) \\
\phi_{17}&=(1.152\sin(x_0),         -70.56\cos(x_1)) \\
\phi_{18}&=(0.710,         -70.56\cos(x_1)) \\
\phi_{19}&=(0.99\cos(x_0),  \\
&    -1.0x_0 - 39.53\sin(x_1) - 152.57\cos(x_1)) \\
\phi_{20}&=(-0.871\sin(x_1),    -3.826\sin(x_0) - 0.211\sin(x_1) + 4.3117) \\
\end{aligned}
\end{equation*}    
\caption{The given set of best-predicted ODEs for Table~\ref{tab:diff-active}.}
\label{fig:eq-list}
\end{figure*}

The Kendall tau distance is computed with two ranked lists, one of them is evaluated on full data and another is evaluated on the chosen region with high uncertainty. We use library\footnote{\url{https://docs.scipy.org/doc/scipy/reference/generated/scipy.stats.kendalltau.html}} to compute the ranking score.

\subsection{Dataset} \label{apx:exp-phase}
We present the visualized phase portraits and the explicit forms of the ODEs considered in this research in the following figures and tables.

\begin{figure*}
    \centering
    \includegraphics[width=0.2\linewidth]{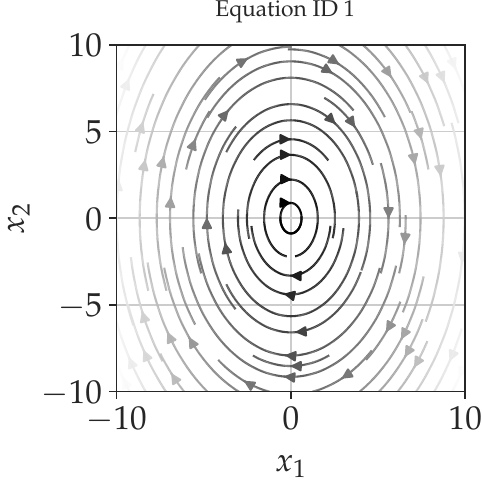}
    \includegraphics[width=0.2\linewidth]{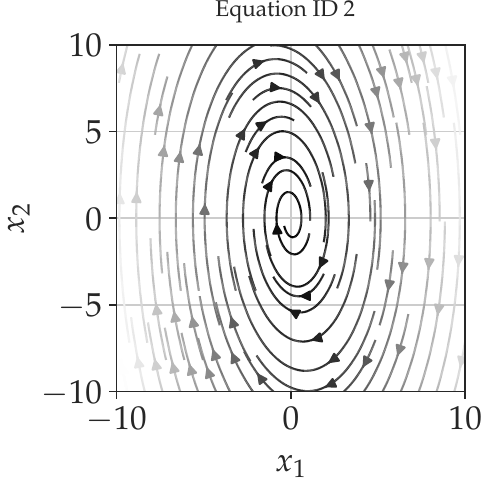}
    \includegraphics[width=0.18\linewidth]{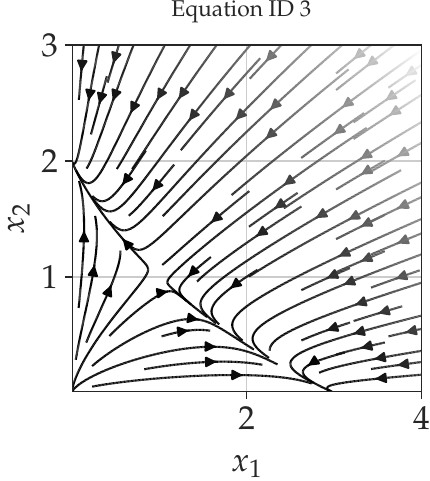}
    \includegraphics[width=0.2\linewidth]{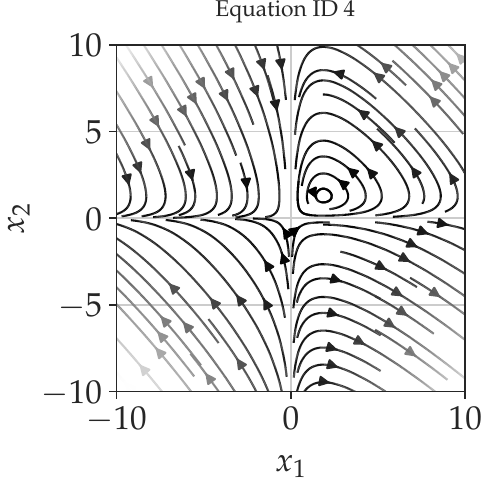}
    \includegraphics[width=0.2\linewidth]{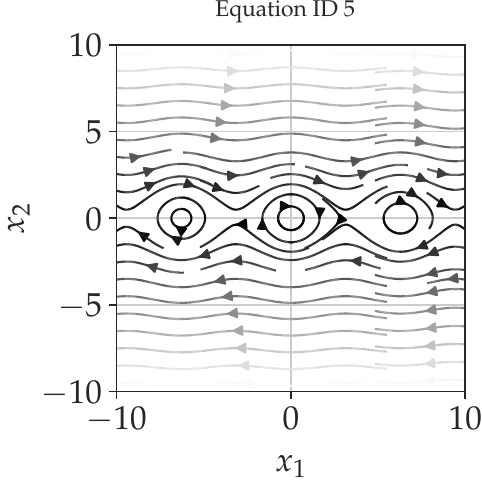}
    \includegraphics[width=0.2\linewidth]{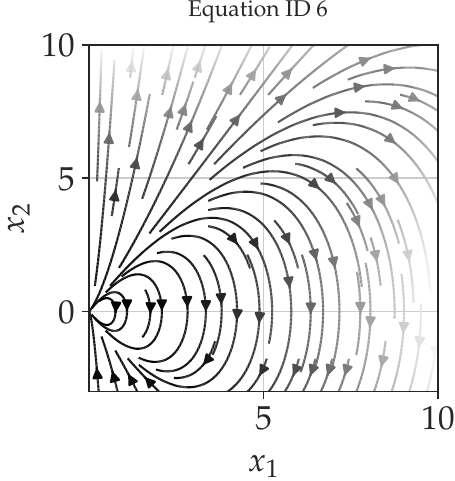}
    \includegraphics[width=0.2\linewidth]{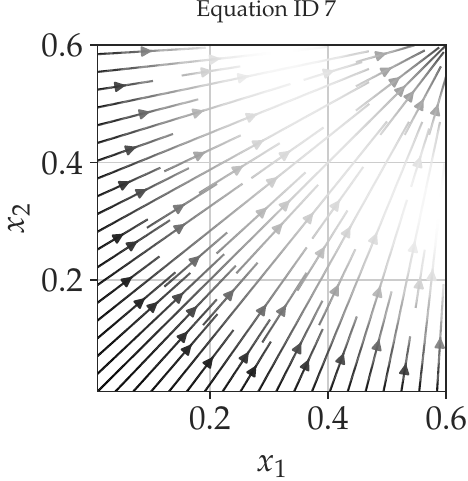}
    \includegraphics[width=0.2\linewidth]{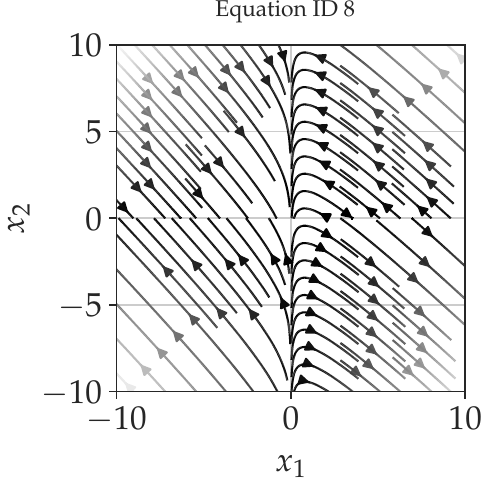}
    \includegraphics[width=0.2\linewidth]{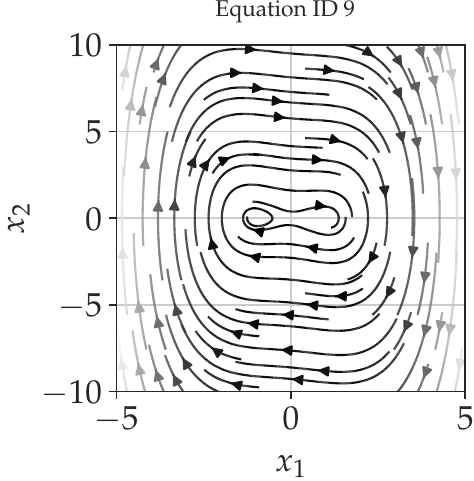}
    \includegraphics[width=0.2\linewidth]{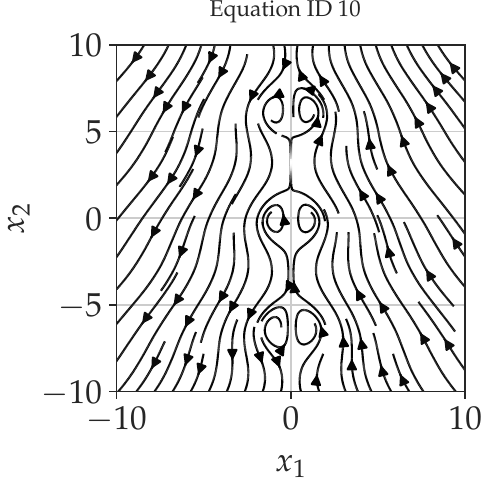}
    \includegraphics[width=0.2\linewidth]{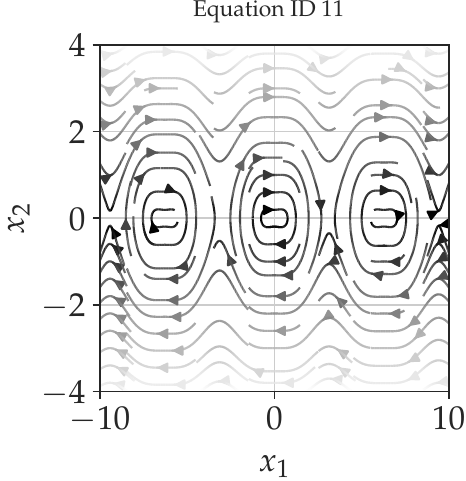}
    \includegraphics[width=0.2\linewidth]{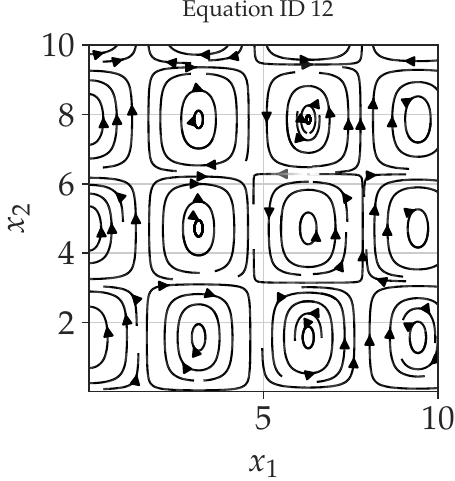}
    \caption{Selected phase portrait of Strogatz dataset with variables $n=2$.}
    \label{fig:strogatz-pp-var2.}
\end{figure*}

\begin{figure*}
    \centering
    \includegraphics[width=0.2\linewidth]{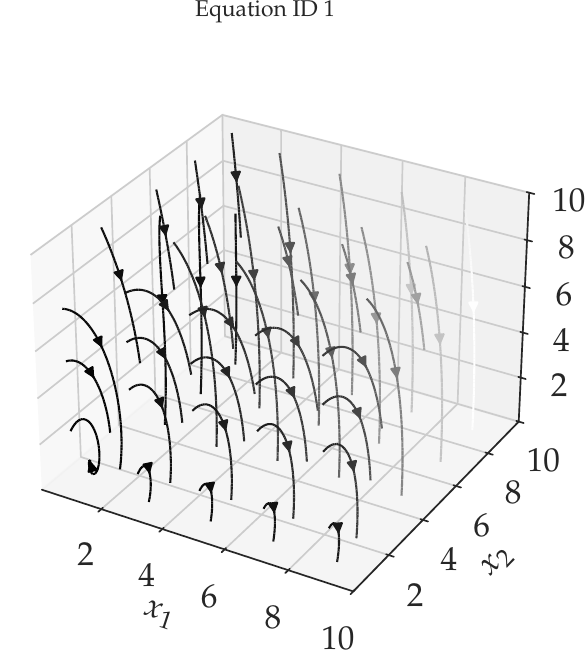}
    \includegraphics[width=0.2\linewidth]{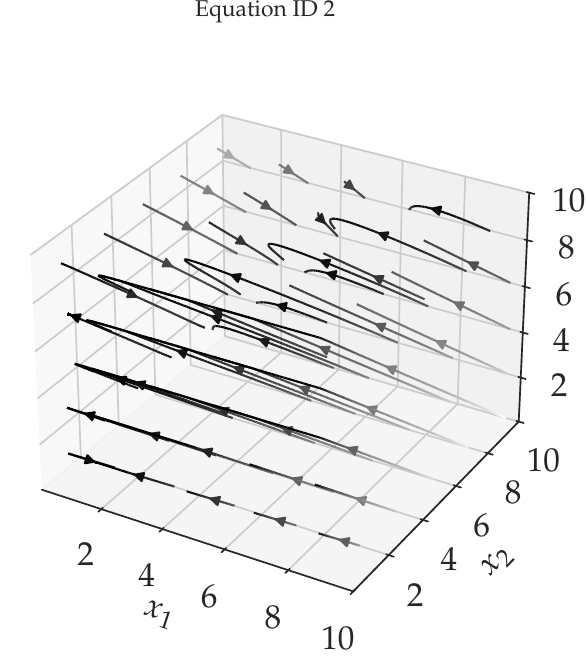}
    \includegraphics[width=0.2\linewidth]{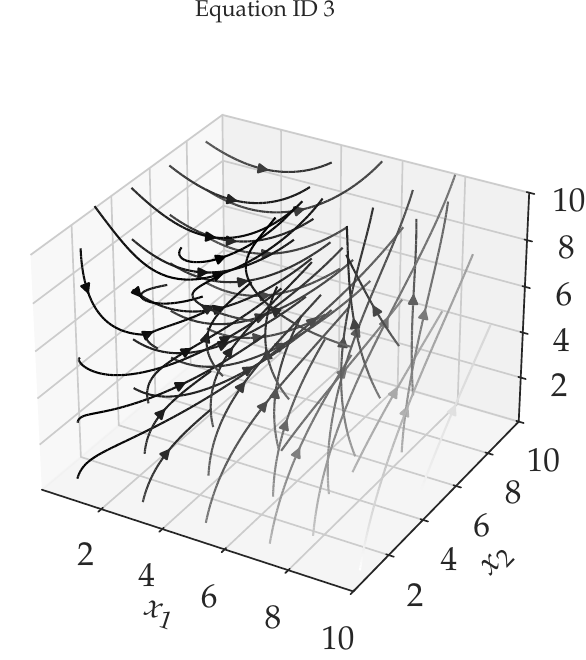}
    \includegraphics[width=0.2\linewidth]{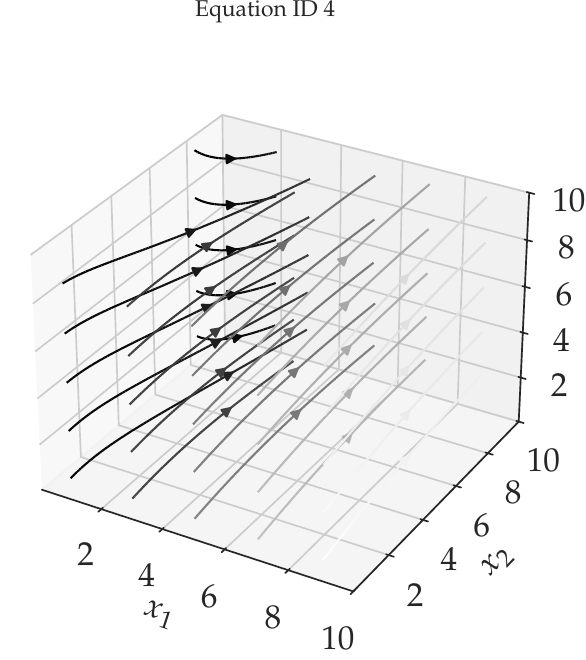}
    \includegraphics[width=0.2\linewidth]{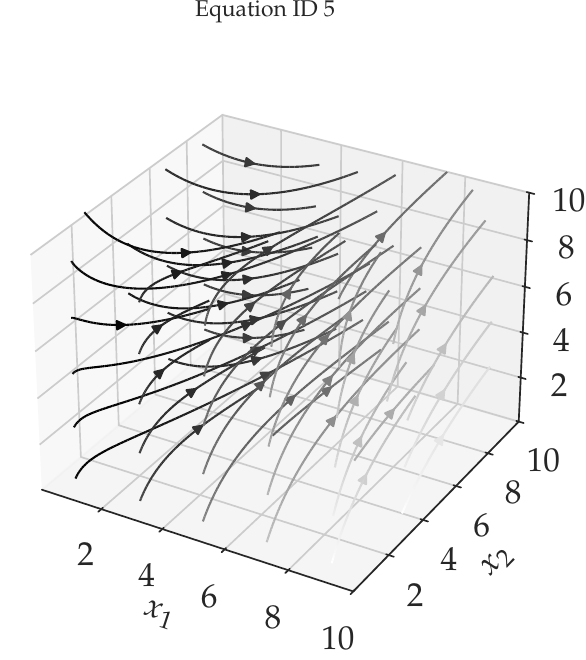}
    \includegraphics[width=0.2\linewidth]{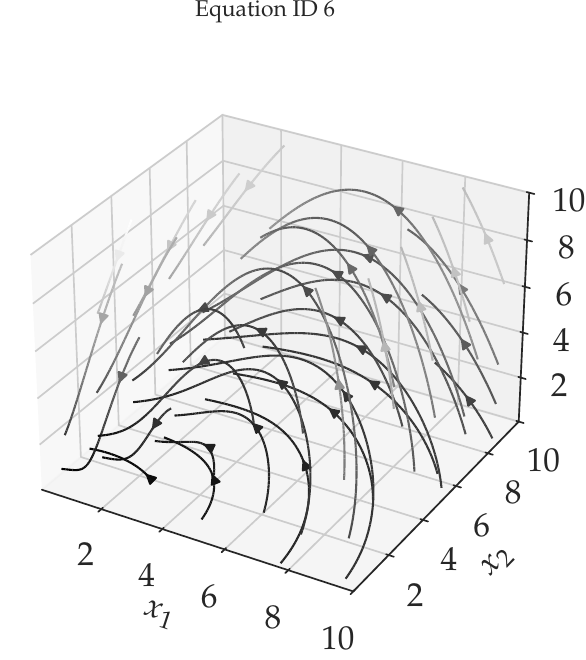}
    \includegraphics[width=0.2\linewidth]{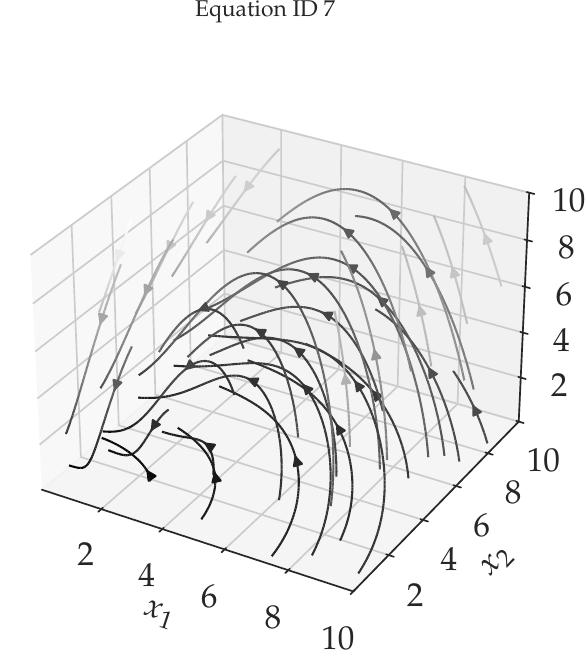}
    \includegraphics[width=0.2\linewidth]{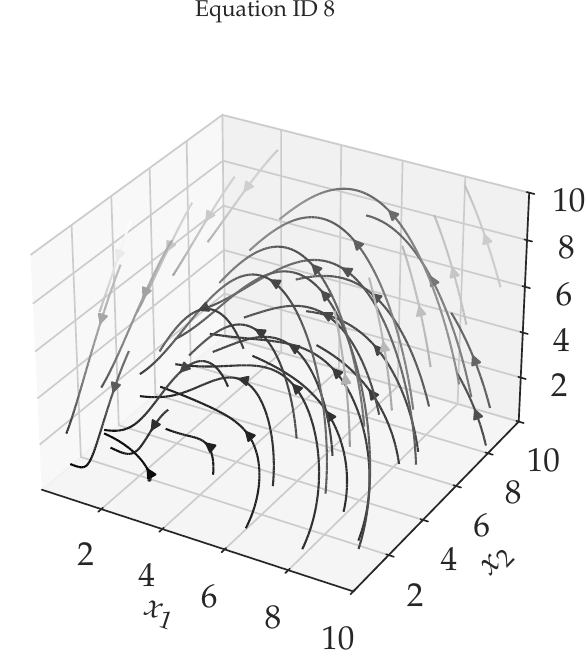}
    \includegraphics[width=0.2\linewidth]{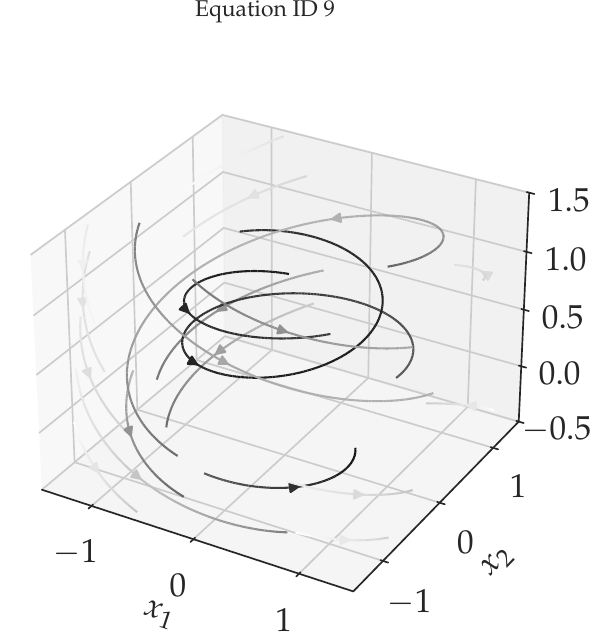}
    \includegraphics[width=0.2\linewidth]{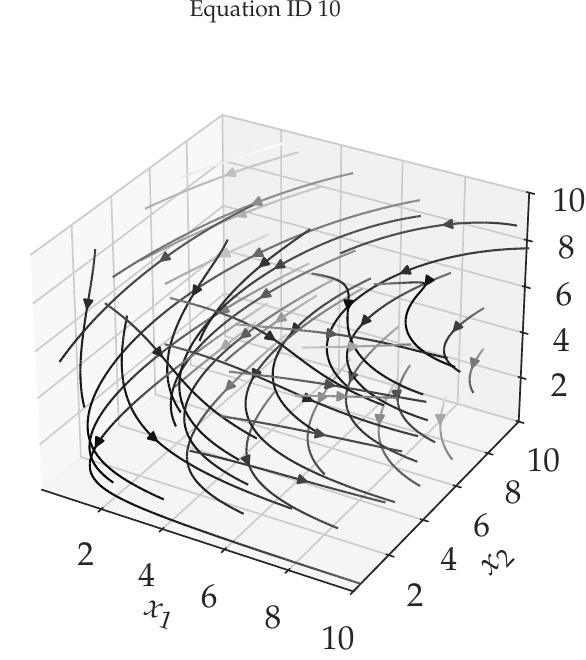}
    \caption{Selected phase portrait of Strogatz dataset with variables $n=3$.}
    \label{fig:strogatz-pp-var3.}
\end{figure*}



\begin{table*}[!ht]
    \centering
    \begin{tabular}{c|l}
    \toprule
        ID &  Equation \\
        \midrule
1  &  RC-circuit (charging capacitor) \\
 & $\dot{x}_0 = (0.7 - x_0 / 1.2) / 2.31$   \\ \hline
{2} &  Population growth (naive) \\
 & $\dot{x}_0 = 0.2{3}x_0$   \\ \hline
{3} &  Population growth with carrying capacity \\
 & $\dot{x}_0 = 0.79x_0(1 - x_0 / 74.3)$   \\ \hline
4  &  RC-circuit with non-linear resistor (charging capacitor) \\
 & $\dot{x}_0 = 1 / (1 + \exp(0.5 - x_0 / 0.96)) - 0.5$   \\ \hline
5  &  Velocity of a falling object with air resistance \\
 & $\dot{x}_0 = 9.81 - 0.0021175x_0^2$   \\ \hline
6  &  Autocatalysis with one fixed abundant chemical \\
 & $\dot{x}_0 = 2.1x_0 - 0.5x_0^2$   \\ \hline
7  &  Gompertz law for tumor growth \\
 & $\dot{x}_0 = 0.03{2}x_0\log(2.29x_0)$   \\ \hline
8  &  Logistic equation with Allee effect \\
 & $\dot{x}_0 = 0.14x_0(1 - x_0 / 130.0)(x_0 / 4.4 - 1)$   \\ \hline
9  &  Language death model for two languages \\
 & $\dot{x}_0 = (1 - x_0)0.3{2}- x_00.28$   \\ \hline
10  &  Refined language death model for two languages \\
 & $\dot{x}_0 = (1 - x_0)0.{2}x_0^{1.2} - x_0(1 - 0.2)(1 - x_0)^{1.2}$   \\ \hline
11  &  Naive critical slowing down (statistical mechanics) \\
 & $\dot{x}_0 = - x_0^3$   \\ \hline
1{2} &  Photons in a laser (simple) \\
 & $\dot{x}_0 = 1.8x_0 - 0.1107x_0^2$   \\ \hline
1{3} &  Overdamped bead on a rotating hoop \\
 & $\dot{x}_0 = 0.0981\sin(x_0)(9.7\cos(x_0) - 1)$   \\ \hline
14  &  Budworm outbreak model with predation \\
 & $\dot{x}_0 = 0.78x_0(1 - x_0 / 81.0) - 0.9x_0^{2}/ (21.2^{2}+ x_0^2)$   \\ \hline
15  &  Budworm outbreak with predation (dimensionless) \\
 & $\dot{x}_0 = 0.4x_0(1 - x_0 / 95.0) - x_0^{2}/ (1 + x_0^2)$   \\ \hline
16  &  Landau equation (typical time scale tau = 1) \\
 & $\dot{x}_0 = 0.1x_0 - -0.04x_0^{3}- 0.001x_0^5$   \\ \hline
17  &  Logistic equation with harvesting/fishing \\
 & $\dot{x}_0 = 0.4x_0(1 - x_0 / 100.0) - 0.3$   \\ \hline
18  &  Improved logistic equation with harvesting/fishing \\
 & $\dot{x}_0 = 0.4x_0(1 - x_0 / 100.0) - 0.24x_0 / (50.0 + x_0)$   \\ \hline
19  &  Improved logistic equation with harvesting/fishing (dimensionless) \\
 & $\dot{x}_0 = x_0(1 - x_0) - 0.08x_0 / (0.8 + x_0)$   \\ \hline
20  &  Autocatalytic gene switching (dimensionless) \\
 & $\dot{x}_0 = 0.1 - 0.55x_0 + x_0^{2}/ (1 + x_0^2)$   \\ \hline
21  &  Dimensionally reduced SIR infection model for dead people (dimensionless) \\
 & $\dot{x}_0 = 1.{2}- 0.{2}x_0 - \exp(-x_0)$   \\ 
    \bottomrule
    \end{tabular}
    \caption{Selected Strogatz dataset with variable $n=1$.}
    \label{tab:odebench-1d}
\end{table*}

\begin{table*}[!h]
    \centering
    \begin{tabular}{c|ll}
    \toprule
        ID & \multicolumn{2}{l}{ Explicit Equations}   \\
        \midrule
1  & \multicolumn{2}{l}{ Harmonic oscillator without damping} \\
 & $\dot{x}_0 = x_1$   & $\dot{x}_1 = - 2.1x_0$   \\ \hline
{2} & \multicolumn{2}{l}{ Harmonic oscillator with damping} \\
 & $\dot{x}_0 = x_1$   & $\dot{x}_1 = - 4.5x_0 - 0.4{3}x_1$   \\ \hline
{3} & \multicolumn{2}{l}{ Lotka-Volterra competition model (Strogatz version with sheeps and rabbits)} \\
 & $\dot{x}_0 = x_0(3.0 - x_0 - 2.0x_1)$   & $\dot{x}_1 = x_1(2.0 - x_0 - x_1)$   \\ \hline
4  & \multicolumn{2}{l}{ Lotka-Volterra simple (as on Wikipedia)} \\
 & $\dot{x}_0 = x_0(1.84 - 1.45x_1)$   & $\dot{x}_1 = - x_1(3.0 - 1.6{2}x_0)$   \\ \hline
5  & \multicolumn{2}{l}{ Pendulum without friction} \\
 & $\dot{x}_0 = x_1$   & $\dot{x}_1 = - 0.9\sin(x_0)$   \\ \hline
6  & \multicolumn{2}{l}{ Dipole fixed point} \\
 & $\dot{x}_0 = 0.65x_0x_1$   & $\dot{x}_1 = x_1^{2}- x_0^2$   \\ \hline
7  & \multicolumn{2}{l}{ RNA molecules catalyzing each others replication} \\
 & $\dot{x}_0 = x_0(x_1 - 1.61x_0x_1)$   & $\dot{x}_1 = x_1(x_0 - 1.61x_0x_1)$   \\ \hline
8  & \multicolumn{2}{l}{ SIR infection model only for healthy and sick} \\
 & $\dot{x}_0 = - 0.4x_0x_1$   & $\dot{x}_1 = 0.4x_0x_1 - 0.314x_1$   \\ \hline
9  & \multicolumn{2}{l}{ Damped double well oscillator} \\
 & $\dot{x}_0 = x_1$   & $\dot{x}_1 = - 0.18x_1 + x_0 - x_0^3$   \\ \hline
10  & \multicolumn{2}{l}{ Glider (dimensionless)} \\
 & $\dot{x}_0 = - \sin(x_1) - 0.08x_0^2$   & $\dot{x}_1 = x_0 - \cos(x_1) / x_0$   \\ \hline
11  & \multicolumn{2}{l}{ Frictionless bead on a rotating hoop (dimensionless)} \\
 & $\dot{x}_0 = x_1$   & $\dot{x}_1 = \sin(x_0)(\cos(x_0) - 0.93)$   \\ \hline
1{2} & \multicolumn{2}{l}{ Rotational dynamics of an object in a shear flow} \\
 & $\dot{x}_0 = \cot(x_1)\cos(x_0)$   & $\dot{x}_1 = \sin(x_0)(\cos(x_1)^{2}+ 4.{2}\sin(x_1)^2)$   \\ \hline
1{3} & \multicolumn{2}{l}{ Pendulum with non-linear damping, no driving (dimensionless)} \\
 & $\dot{x}_0 = x_1$   & $\dot{x}_1 = - \sin(x_0) - x_1 - 0.07\cos(x_0)x_1$   \\ \hline
14  & \multicolumn{2}{l}{ Van der Pol oscillator (standard form)} \\
 & $\dot{x}_0 = x_1$   & $\dot{x}_1 = - x_0 - 0.4{3}(x_0^{2}- 1)x_1$   \\ \hline
15  & \multicolumn{2}{l}{ Van der Pol oscillator (simplified form from Strogatz)} \\
 & $\dot{x}_0 = 3.37(x_1 - x_0^{3}/ {3}+ x_0)$   & $\dot{x}_1 = - x_0 / 3.37$   \\ \hline
16  & \multicolumn{2}{l}{ Glycolytic oscillator, e.g., ADP and F6P in yeast (dimensionless)} \\
 & $\dot{x}_0 = - x_0 + 2.4x_1 + x_0^{2}x_1$   & $\dot{x}_1 = 0.07 - 2.4x_0 - x_0^{2}x_1$   \\ \hline
17  & \multicolumn{2}{l}{ Duffing equation (weakly non-linear oscillation)} \\
 & $\dot{x}_0 = x_1$   & $\dot{x}_1 = - x_0 + 0.886x_1(1 - x_0^2)$   \\ \hline
18  & \multicolumn{2}{l}{ Cell cycle model by Tyson for interaction between protein cdc{2}and cyclin (dimensionless)} \\
 & $\dot{x}_0 = 15.{3}(x_1 - x_0)(0.001 + x_0^2) - x_0$   & $\dot{x}_1 = 0.{3}- x_0$   \\ \hline
19  & \multicolumn{2}{l}{ Reduced model for chlorine dioxide-iodine-malonic acid rection (dimensionless)} \\
 & $\dot{x}_0 = 8.9 - x_0 - 4.0x_0x_1 / (1 + x_0^2)$   & $\dot{x}_1 = 1.4x_0(1 - x_1 / (1 + x_0^2))$   \\ \hline
20  & \multicolumn{2}{l}{ Driven pendulum with linear damping / Josephson junction (dimensionless)} \\
 & $\dot{x}_0 = x_1$   & $\dot{x}_1 = 1.67 - \sin(x_0) - 0.64x_1$   \\ \hline
21  & \multicolumn{2}{l}{ Driven pendulum with quadratic damping (dimensionless)} \\
 & $\dot{x}_0 = x_1$   & $\dot{x}_1 = 1.67 - \sin(x_0) - 0.64x_1^1 $   \\
    \bottomrule
    \end{tabular}
    \caption{Selected Strongatz dataset with variables $n=2$.}
    \label{tab:odebench-2d}
\end{table*}

\begin{table*}[!ht]
    \centering
    \begin{tabular}{c|l}
    \toprule
        ID & Explicit Equations \\
        \midrule
1  & { Maxwell-Bloch equations (laser dynamics)} \\
 & $\dot{x}_0 = 0.1(x_1 - x_0)$  \\
 & $\dot{x}_1 = 0.21(x_0x_{2}- x_1)$   \\
 & $\dot{x}_{2}= 0.34(3.1 + 1 - x_{2}- 3.1x_0x_1)$   \\ \hline
{2} & { Model for apoptosis (cell death)} \\
 & $\dot{x}_0 = 0.1 - 0.4x_1x_0 / (0.1 + x_0) - 0.05x_0$\\
 & $\dot{x}_1 = 0.6x_{2}(0.1 + x_1) - 0.{2}x_1 / (0.1 + x_1) - 7.95x_0x_1 / (2.0 + x_1)$   \\
 & $\dot{x}_{2}= - 0.6x_{2}(0.1 + x_1) + 0.{2}x_1 / (0.1 + x_1) + 7.95x_0x_1 / (2.0 + x_1)$   \\ \hline
{3} & { Lorenz equations in well-behaved periodic regime} \\
 & $\dot{x}_0 = 5.1(x_1 - x_0)$  \\
 & $\dot{x}_1 = 12.0x_0 - x_1 - x_0x_2$  \\
 & $\dot{x}_{2}= x_0x_1 - 1.67x_2$   \\ \hline
4  & { Lorenz equations in complex periodic regime} \\
 & $\dot{x}_0 = 10.0(x_1 - x_0)$   \\
 & $\dot{x}_1 = 99.96x_0 - x_1 - x_0x_2$   \\
 & $\dot{x}_{2}= x_0x_1 - 2.6666666666666665x_2$   \\ \hline
5  & { Lorenz equations standard parameters (chaotic)} \\
 & $\dot{x}_0 = 10.0(x_1 - x_0)$   \\
 & $\dot{x}_1 = 28.0x_0 - x_1 - x_0x_2$   \\
 & $\dot{x}_{2}= x_0x_1 - 2.6666666666666665x_2$   \\ \hline
6  & { Rössler attractor (stable fixed point)} \\
 & $\dot{x}_0 = 5.0(- x_1 - x_2)$  \\
 & $\dot{x}_1 = 5.0(x_0 -0.{2}x_1)$   \\& $\dot{x}_{2}= 5.0(0.{2}+ x_{2}(x_0 - 5.7))$   \\ \hline
7  & { Rössler attractor (periodic)} \\
 & $\dot{x}_0 = 5.0(- x_1 - x_2)$   \\
 & $\dot{x}_1 = 5.0(x_0  + 0.1x_1)$   \\
 & $\dot{x}_{2}= 5.0(0.{2}+ x_{2}(x_0 - 5.7))$   \\ \hline
8  & { Rössler attractor (chaotic)} \\
 & $\dot{x}_0 = 5.0(- x_1 - x_2)$   \\
 & $\dot{x}_1 = 5.0(x_0  + 0.{2}x_1)$  \\ & $\dot{x}_{2}= 5.0(0.{2}+ x_{2}(x_0 - 5.7))$   \\ \hline
9  & { Aizawa attractor (chaotic)} \\
 & $\dot{x}_0 = x_0(x_{2}- 0.7) - 3.5x_1$  \\
 & $\dot{x}_1 = 3.5x_0 + x_1(x_{2}- 0.7)$   \\
 & $\dot{x}_{2}= 0.65 + 0.95x_{2}- x_2^{3}/ 3. - (x_0^{2}+ x_1^2)(1 + 0.25x_2) + 0.1x_{2}x_0^3$   \\ \hline
10  & { Chen-Lee attractor} \\
 & $\dot{x}_0 = 5x_0 - x_1x_2$   \\
 & $\dot{x}_1 = -10.0x_1 + x_0x_2$  \\
 & $\dot{x}_{2}= -3.8x_{2}+ x_0x_1 / 3.0$ \\
    \bottomrule
    \end{tabular}
    \caption{The Strogatz dataset with variables $n=3$.}
    \label{tab:odebench-3d}
\end{table*}

\begin{table*}[!ht]
    \centering
    \begin{tabular}{c|lll}
    \toprule
        ID &  Equations \\
        \midrule
1  & \multicolumn{2}{l}{ Norel1990 - MPF and Cyclin Oscillations} \\
 & $\dot{x}_0 = 1.0x_0^2x_1 - 10.0x_0/(x_0 + 1.0) + 3.466x_1$ \\
 & $\dot{x}_1 = 1.2 - 1.0x_0$ \\
  \hline
2  & \multicolumn{2}{l}{ Chrobak2011 - A mathematical model of induced cancer-adaptive immune system competition} \\
 & $\dot{x}_0 = -0.03125x_0^2 - 0.125x_0x_1 + 0.0625x_0$ \\
 & $\dot{x}_1 = -0.08594x_0x_1 - 0.03125x_1^2 + 0.03125x_1$ \\
  \hline
3  & \multicolumn{2}{l}{ FitzHugh1961-NerveMembrane} \\
 & $\dot{x}_0 = -1.0x_0^3 + 3.0x_0 + 3.0x_1 - 1.2$ \\
 & $\dot{x}_1 = -0.3333x_0 - 0.2667x_1 + 0.2333$ \\
  \hline
4  & \multicolumn{2}{l}{ Clarke2000 - One-hit model of cell death in neuronal degenerations} \\
 & $\dot{x}_0 = -0.278x_0$ \\
 & $\dot{x}_1 = -0.223x_1$ \\
  \hline
5  & \multicolumn{2}{l}{ Wodarz2018/1 - simple model} \\
 & $\dot{x}_0 = 0.004x_0 + 0.004x_1/(0.01x_0^1.0 + 1.0)$ \\
 & $\dot{x}_1 = 0.006x_0 - 0.003x_1 - 0.004x_1/(0.01x_0^1.0 + 1.0)$ \\
  \hline
6  & \multicolumn{2}{l}{ Ehrenstein2000 - Positive-Feedback model for the loss of acetylcholine in Alzheimer's disease} \\
 & $\dot{x}_0 = -0.007x_0x_1$ \\
 & $\dot{x}_1 = -0.004x_0 - 0.01x_1 + 0.33$ \\
  \hline
7  & \multicolumn{2}{l}{ Cao2013 - Application of ABSIS method in the bistable Schlagl model} \\
 & $\dot{x}_0 = 0.12x_0^2 - 3.071x_0 + 12.5 - 0.00192/x_0$ \\
 & $\dot{x}_1 = -0.12x_0^2 + 3.071x_0 - 12.5 + 0.00192/x_0$ \\
  \hline
8  & \multicolumn{2}{l}{ Chaudhury2020 - Lotka-Volterra mathematical model of CAR-T cell and tumour kinetics} \\
 & $\dot{x}_0 = 0.002x_0x_1 - 0.16x_0$ \\
 & $\dot{x}_1 = 0.15x_1$ \\
  \hline
9  & \multicolumn{2}{l}{ Baker2013 - Cytokine Mediated Inflammation in Rheumatoid Arthritis} \\
 & $\dot{x}_0 = -x_0 + 3.5x_1^2/(x_1^2 + 0.25)$ \\
 & $\dot{x}_1 = 1.0x_1^2/(x_0^2x_1^2 + x_0^2 + x_1^2 + 1.0) - 1.25x_1 + 0.025/(x_0^2 + 1.0)$ \\
  \hline
10  & \multicolumn{2}{l}{ Somogyi1990-CaOscillations} \\
 & $\dot{x}_0 = -5.0x_0x_1^4.0/(x_1^4.0 + 81.0) - 0.01x_0 + 2.0x_1$ \\
 & $\dot{x}_1 = 5.0x_0x_1^4.0/(x_1^4.0 + 81.0) + 0.01x_0 - 3.0x_1 + 1.0$ \\
  \hline
11  & \multicolumn{2}{l}{ Cucuianu2010 - A hypothetical-mathematical model of acute myeloid leukaemia pathogenesis} \\
 & $\dot{x}_0 = -0.1x_0 + 0.3x_0/(0.5x_0 + 0.5x_1 + 1.0)$ \\
 & $\dot{x}_1 = -0.1x_1 + 0.3x_1/(0.5x_0 + 0.5x_1 + 1.0)$ \\
  \hline
12  & \multicolumn{2}{l}{ Wang2016/3 - oncolytic efficacy of M1 virus-SN model} \\
 & $\dot{x}_0 = -0.2x_0x_1 - 0.02x_0 + 0.02$ \\
 & $\dot{x}_1 = 0.16x_0x_1 - 0.03x_1$ \\
  \hline
13  & \multicolumn{2}{l}{ Chen2011/1 - bone marrow invasion absolute model} \\
 & $\dot{x}_0 = -0.2x_0^2 + 0.1x_0$ \\
 & $\dot{x}_1 = -1.0x_0x_1 - 0.8x_1^2 + 0.7x_1$ \\
  \hline
14  & \multicolumn{2}{l}{ Cao2013 - Application of ABSIS method in the reversible isomerization model} \\
 & $\dot{x}_0 = -0.12x_0 + 1.0x_1$ \\
 & $\dot{x}_1 = 0.12x_0 - 1.0x_1$ \\
    \bottomrule
    \end{tabular}
    \caption{Selected ODEBase dataset with variables $n=2$.}
    \label{tab:odebase-2d}
\end{table*}

\begin{table*}[!ht]
    \centering
    \begin{tabular}{c|l}
    \toprule
1  & { Turner2015-Human/Mosquito ELP Model} \\
 & $\dot{x}_0 = 600.0 - 0.411x_0$ \\
 & $\dot{x}_1 = 0.361x_0 - 0.184x_1$ \\
 & $\dot{x}_2 = 0.134x_1 - 0.345x_2$ \\
  \hline
2  & { Al-Tuwairqi2020 - Dynamics of cancer virotherapy - Phase I treatment} \\
 & $\dot{x}_0 = -1.0x_0x_2$ \\
 & $\dot{x}_1 = 1.0x_0x_2 - 1.0x_1$ \\
 & $\dot{x}_2 = -0.02x_0x_2 + 1.0x_1 - 0.15x_2$ \\
  \hline
3  & { Fassoni2019 - Oncogenesis encompassing mutations and genetic instability} \\
 & $\dot{x}_0 = 0.01 - 0.01x_0$ \\
 & $\dot{x}_1 = 0.03x_1$ \\
 & $\dot{x}_2 = -0.5x_2^2 + 0.034x_2$ \\
  \hline
4  & { Zatorsky2006-p53-Model5} \\
 & $\dot{x}_0 = -3.7x_0x_1 + 2.0x_0$ \\
 & $\dot{x}_1 = -0.9x_1 + 1.1x_2$ \\
 & $\dot{x}_2 = 1.5x_0 - 1.1x_2$ \\
  \hline
5  & { Lenbury2001-InsulinKineticsModel-A} \\
 & $\dot{x}_0 = -0.1x_0x_2 + 0.2x_1x_2 + 0.1x_2$ \\
 & $\dot{x}_1 = -0.01x_0 + 0.01 + 0.01/x_2$ \\
 & $\dot{x}_2 = -0.1x_1x_2 + 0.257x_1 - 0.1x_2^2 + 0.331x_2 - 0.3187$ \\
  \hline
6  & { Zatorsky2006-p53-Model1} \\
 & $\dot{x}_0 = -3.2x_0x_1 + 0.3$ \\
 & $\dot{x}_1 = -0.1x_1 + 0.1x_2$ \\
 & $\dot{x}_2 = 0.4x_0 - 0.1x_2$ \\
  \hline
7  & { Smallbone2013 - Colon Crypt cycle - Version 1} \\
 & $\dot{x}_0 = -0.002207x_0^2 - 0.002207x_0x_1 - 0.002207x_0x_2 + 0.1648x_0$ \\
 & \footnotesize{$\dot{x}_1 = -0.01312x_0^2 - 0.0216x_0x_1 - 0.01312x_0x_2 + 1.574x_0 - 0.008477x_1^2 - 0.008477x_1x_2 + 0.5972x_1$} \\
 & $\dot{x}_2 = -0.04052x_0x_1 - 0.04052x_1^2 - 0.04052x_1x_2 + 4.863x_1 - 1.101x_2$ \\
  \hline
8  & { Cortes2019 - Optimality of the spontaneous prophage induction rate.} \\
 & $\dot{x}_0 = -0.99x_0^2/(x_0 + x_1) - 1.0x_0x_1/(x_0 + x_1) + 0.99x_0$ \\
 & $\dot{x}_1 = -0.99x_0x_1/(x_0 + x_1) - 1.0x_1^2/(x_0 + x_1) + 1.0x_1$ \\
 & $\dot{x}_2 = -0.001x_2$ \\
  \hline
9  & { Figueredo2013/2 - immunointeraction model with IL2} \\
 & $\dot{x}_0 = -1.0x_0x_1/(x_0 + 1.0) + 0.18x_0$ \\
 & $\dot{x}_1 = 0.05x_0 + 0.124x_1x_2/(x_2 + 20.0) - 0.03x_1$ \\
 & $\dot{x}_2 = 5.0x_0x_1/(x_0 + 10.0) - 10.0x_2$ \\
  \hline
10  & A mathematical model of cytotoxic and helper T cell interactions in a tumour microenvironment \\
 & $\dot{x}_0 = -10.0x_0^2 - 2.075x_0x_2 + 10.0x_0$ \\
 & $\dot{x}_1 = 0.19x_0x_1/(x_0^2 + 0.0016) - 1.0x_1 + 0.5$ \\
 & $\dot{x}_2 = -2.075x_0x_2 + 1.0x_1x_2 - 1.0x_2 + 2.0$ \\
  \hline
11  & { Munz2009 - Zombie SIZRC} \\
 & $\dot{x}_0 = -0.009x_0x_1 + 0.05$ \\
 & $\dot{x}_1 = 0.004x_0x_1$ \\
 & $\dot{x}_2 = 0.005x_0x_1$ \\
  \bottomrule
    \end{tabular}
    \caption{Selected ODEBase dataset with variables $n=3$.}
    \label{tab:odebase-3part1}
\end{table*}